\pdfoutput=1

\documentclass[10pt,twocolumn,letterpaper]{article}

\usepackage{cvpr}
\usepackage{times}
\usepackage{epsfig}
\usepackage{graphicx}
\usepackage{amsmath}
\usepackage{amssymb}
\usepackage{enumitem}

\makeatletter
\newcommand{\printfnsymbol}[1]{%
  \textsuperscript{\@fnsymbol{#1}}%
}
\makeatother

\makeatletter
\renewcommand\@makefntext[1]{\leftskip=0em\hskip-0em\@makefnmark#1}
\makeatother

\usepackage[pagebackref=true,breaklinks=true,colorlinks,bookmarks=false]{hyperref} 

\cvprfinalcopy 

\def\cvprPaperID{1910} 

\newcommand{\SNET}{S-NET }
\newcommand{\SNETwo}{S-NET}

\ifcvprfinal\fi
\begin{document}

\title{Learning to Sample}

\author{Oren Dovrat\printfnsymbol{1} \\
Tel-Aviv University\\
{\tt\small Oren.Dovrat@gmail.com}
\and
Itai Lang\printfnsymbol{1} \\
Tel-Aviv University\\
{\tt\small Itai.Lang83@gmail.com}
\and
Shai Avidan\\
Tel-Aviv University\\
{\tt\small avidan@eng.tau.ac.il}
}

\maketitle
\thispagestyle{empty}

\begin{abstract}

Processing large point clouds is a challenging task. Therefore, the data is often sampled to a size that can be processed more easily. The question is how to sample the data? A popular sampling technique is Farthest Point Sampling (FPS). However, FPS is agnostic to a downstream application (classification, retrieval, etc.). The underlying assumption seems to be that minimizing the farthest point distance, as done by FPS, is a good proxy to other objective functions.

We show that it is better to learn how to sample. To do that, we propose a deep network to simplify 3D point clouds. The network, termed S-NET, takes a point cloud and produces a smaller point cloud that is optimized for a particular task. The simplified point cloud is not guaranteed to be a subset of the original point cloud. Therefore, we match it to a subset of the original points in a post-processing step. We contrast our approach with FPS by experimenting on two standard data sets and show significantly better results for a variety of applications. Our code is publicly available\footnote {\url{https://github.com/orendv/learning_to_sample} \\ \printfnsymbol{1}Equal contribution}



\end{abstract}

\section{Introduction}
Capturing 3D data is getting easier in recent years and there is a growing number of 3D shape repositories available online. This data can be represented in a variety of ways, including point clouds, multi-view images and voxel grids. A point cloud contains information only about the surface of a 3D object, while a grid based representation also holds data about free space, making the former much more efficient. However, processing a point cloud can be challenging, since it may contain a lot of data points. Reducing the number of points can be beneficial in many aspects, such as reduction of power consumption, computational cost and communication load, to name a few.

One naive approach to reduce the data load is to randomly sample a subset of points. Another approach, which is commonly used in the literature, is Farthest Point Sampling (FPS)~\cite{qi2017pointnet, qi2017pointnetplusplus, li2018point_cnn}. This sampling method takes into account the structure of the point cloud and selects a group of points that are farthest apart from each other~\cite{eldar1997FPS, moenning2003fast_marching}. These sampling methods, as well as other approaches in the literature~\cite{alexa2001point_set_surfaces, linsen2001point_cloud_representation}, operate according to a non-learned predetermined rule.

In the last few years, deep learning techniques have been applied with great success to point cloud data. Among various applications one can find point cloud classification~\cite{qi2017pointnet, qi2017pointnetplusplus, li2018point_cnn, zaheer2017deep_sets, shen2018mining_local_structures, wang2018dgcnn}, part segmentation~\cite{qi2017pointnet, qi2017pointnetplusplus, li2018point_cnn, shen2018mining_local_structures, wang2018dgcnn, li2018so_net}, semantic segmentation~\cite{qi2017pointnet, li2018point_cnn, wang2018dgcnn, tchapmi2017seg_cloud, su2018splatnet, hua2018pointwise_cnn} and retrieval~\cite{uy2018pointnetvlad, kuang2018model_retrieval}. Other techniques perform point cloud auto-encoding ~\cite{achlioptas2018latent_pc, yang2018folding_net, groueix2018atlas_net}, generation~\cite{achlioptas2018latent_pc, sun2018point_grow, li2018pointcloudgan}, completion~\cite{achlioptas2018latent_pc, agrawal2018semantic_shape_completion, yuan2018pcn} and up-sampling~\cite{yu2018pu_net, yu2018ec_net, zhang2018upsampling_pc}. Yet a learned point clouds sampling approach, subject to a subsequent task objective, has not been proposed before.

\begin{figure}
\begin{center}
\includegraphics[width=1.0\linewidth]{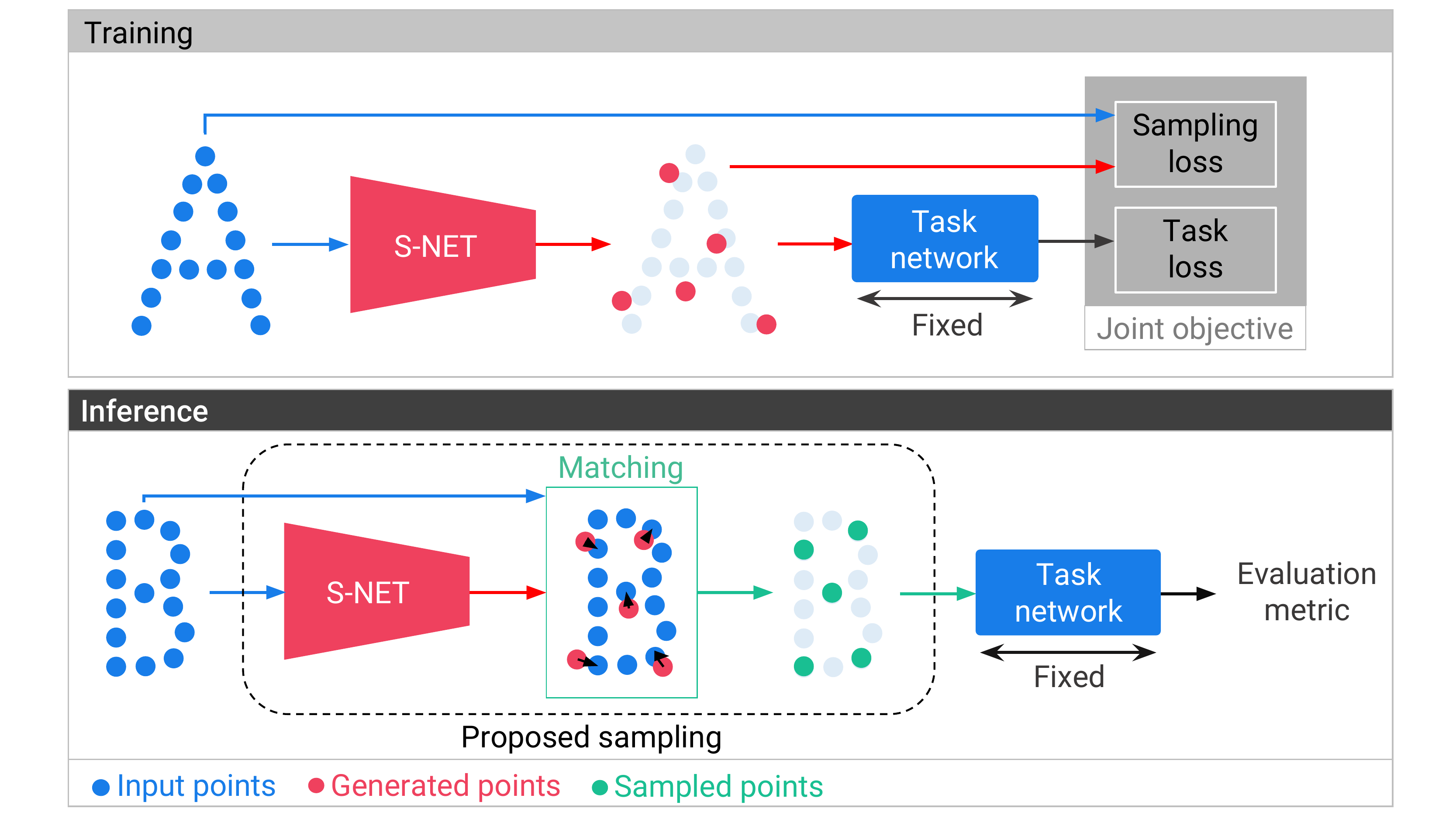}
\end{center}
\caption{{\bfseries An illustration of the proposed learned sampling approach.} In the {\em training} phase, \SNET generates points that are passed to a task network, which was pre-trained and is held fixed. The minimization objective contains the task's loss and a sampling loss. The latter serves as a regularizer and encourages proximity between the input and generated points. At {\em inference} time, we match the points generated by \SNET with the input point cloud and get a subset of it. Only these points are then fed to the task network for performance evaluation.}
\label{fig:overall_method}
\end{figure}



We propose a simplification network, termed \SNETwo, that is based on the architecture of PointNet \cite{qi2017pointnet}. \SNET learns to generate a smaller (simplified) point cloud that is optimized for a downstream task, such as classification, retrieval or reconstruction.

The simplified point cloud must balance two conflicting constraints. On the one hand, we would like it to preserve similarity to the original shape. On the other hand, we wish to optimize it to a subsequent task. We solve this by training the network to generate a set of points that satisfy two objectives: a sampling loss and the task's loss. The sampling loss drives the generated points close to the input point cloud. The task loss ensures that the points are optimal for the task.

An advantages of FPS is that it samples a subset of the original points. In contrast, the simplified point cloud produced by \SNET is not guaranteed to be a subset of the input point cloud. To address this issue, we perform a post-processing step at {\em inference} time, where we match the generated points with the input point cloud and obtain a subset of it, i.e., a set of sampled points (see Figure~\ref{fig:overall_method}). Experiments show that better results for several tasks are achieved using our sampled points in comparison to FPS. 

Our approach can be thought of as a feature selection mechanism~\cite{guyon2003feature_selection, kohavi1997feature_subset_selection}. Each point is a feature of the underlying shape and we seek to select the ones that contribute the most to the task. It can also be interpreted as a form of visual attention~\cite{mnih2014visual_attention, ba2015visual_attention}, focusing the subsequent task network on the significant points.

\SNET is trained to output a fixed sample size, which means that we need to train a different \SNET for every target size. To overcome this limitation, we introduce an extension of \SNET termed ProgressiveNet. ProgressiveNet orders points by importance to the task. This lets the sample size to be chosen at inference time, allowing for a dynamic level-of-detail management, according to the requirements and available resources. 


The proposed sampling approach is applied to three different tasks: point cloud classification, retrieval and reconstruction. We compare our approach with common non-data driven methods: random sampling and FPS. For the first task we show better classification accuracy; in the second we show improved retrieval results; and in the last we get a lower reconstruction error.
To summarize, our key contributions are:
\begin{itemize}[noitemsep, nolistsep]
    \item A task-specific data-driven sampling approach for point clouds;
    \item A Progressive sampling method that orders points according to their relevance for the task;
    \item Improved performance for point cloud classification, retrieval and reconstruction with sampled point clouds.
\end{itemize}

\section{Related work}
\noindent {\bfseries Point cloud simplification and sampling}\quad Several techniques for either point cloud simplification~\cite{pauly2002efficient_simplification, moenning2003new_simplification} or sampling~\cite{katz2013visual_comprehension, chen2018fast_resampling} have been proposed in the literature. Pauly \etal~\cite{pauly2002efficient_simplification} presented and analyzed several simplification methods for point-sampled surfaces, including: clustering methods, iterative simplification and particle simulation. The simplified point set, resulting from these algorithms, was not restricted to be a subset of the original one. Farthest point sampling was adopted in the work of Moenning and Dodgson~\cite{moenning2003new_simplification} as a means to simplify point clouds of geometric shapes, in a uniform as well as feature-sensitive manner.

Katz and Tal~\cite{katz2013visual_comprehension} suggested a view dependent algorithm to reduce the number of points. They used hidden-point removal and target-point occlusion operators in order to improve a human comprehension of the sampled point set. Recently, Chen \etal~\cite{chen2018fast_resampling} employed graph-based filters to extract per point features. Points that preserve specific information are likely to be selected by a their sampling strategy. The desired information is assumed to be beneficial to a subsequent application.

The above sampling approaches aim to optimize a variety of sampling objectives. However, they do not consider {\em directly} the objective of the task to be followed.

\medskip
\noindent {\bfseries Progressive simplification}\quad In a seminal paper, Hoppe ~\cite{hoppe1996progressive_meshes} proposed a technique for progressive mesh simplification. In each step of his method, one edge is collapsed such that minimal geometric distortion is introduced.

A recent work by Hanocka \etal~\cite{hanocka2018mesh_cnn} suggested a neural network that performs task-driven mesh simplification. Their network relies on the edges between the mesh vertices. This information is not available for point clouds.

Several researchers studied the topic of point set compression~\cite{peng2003progressive_geometry, huang2006progressive_geometry, schnabel2006point_cloud_compression}. An octree data structure was used for progressive encoding of the point cloud. The objective of the compression process was low distortion error.

\medskip
\noindent {\bfseries Deep learning on point sets}\quad The pioneering work of Qi \etal~\cite{qi2017pointnet} presented PointNet, the first neural network that operates directly on unordered point cloud data. They constructed their network from per-point multi-layer perceptrons, a symmetric pooling operation and several fully connected layers. PointNet was employed for classification and segmentation tasks and showed impressive results. For assessing the applicability of PointNet for reduced number of input points, they used random sampling and FPS. In our work, we suggest a data-driven sampling approach, that improves the classification performance with sampled sets in comparison to these sampling methods.

Later on, Qi \etal extended their network architecture for hierarchical feature learning~\cite{qi2017pointnetplusplus}. In the training phase, centroid points for local feature aggregation were selected by FPS. Similar to their previous work \cite{qi2017pointnet}, FPS was used for evaluating the ability of their network to operate on fewer input points.

Li \etal~\cite{li2018so_net} suggested to learn the centroid points for feature aggregation by a self-organizing map (SOM). They used feature propagation between points and SOM nodes and showed improved results for point cloud classification and part segmentation. The SOM was optimized separately, as a pre-processing step. 

Building on the work of Qi \etal~\cite{qi2017pointnet}, Achlioptas \etal~\cite{achlioptas2018latent_pc} developed autoencoders and generative adversarial networks for point clouds. Instead of shape class or per-point label, the output of their network was a set of 3D points. In this work we apply our sampling approach for the task of point cloud reconstruction with the autoencoder proposed by Achlioptas \etal.

Several researchers tackled the problem of point cloud consolidation. Yu \etal~\cite{yu2018pu_net} extracted point clouds from geodesic patches of mesh models. They randomly sampled the point sets and trained a network to reconstruct the original patch points. Their follow-up work ~\cite{yu2018ec_net} incorporated edge information to improve the reconstruction accuracy. Zhang \etal~\cite{zhang2018upsampling_pc} studied the influence of sampling strategy on point cloud up-sampling. They used Monte-Carlo random sampling and curvature based sampling. Their network was trained to produce a fixed size point cloud from its sample. In contrast to these works, our study focuses on the down-sampling strategy.


\newcommand{\argmin}{\arg\!\min}
\section{Method}
\noindent {\bf Problem statement}\quad Given a point set $P=\{p_i\in\mathbb{R}^3, i = 1, \dots, n\}$, a sample size $k \leq n$ and a task network $T$, find a subset $S^*$ of $k$ points that minimizes the task network's objective function $f$:
\begin{equation} 
S^* = \argmin_S f(T(S)), \quad S \subset P, \quad |S|=k \leq n.
\end{equation}

This problem poses a challenge, as sampling might seems akin to pooling, yet in pooling the pooled value is propagated forward, so the gradient with respect to it can be calculated. Discrete sampling, however, is like "arg-pooling", where the propagated value cannot be updated incrementally. As a result, a sampling operation cannot be trained directly. Therefore, we propose a two-step process: first, we employ a neural network, i.e., \SNETwo, to generate a set of points. Second, we match the {\em generated} points with the input point cloud to obtain a subset of its points, i.e., the {\em sampled} points. Figure~\ref{fig:overall_method} illustrates the process. 

The input to \SNET is a set of $n$ 3D coordinates, namely points, representing a 3D shape. The output of \SNET is $k$ generated points. \SNET is followed by a task network. The task network is pre-trained on an input of $n$ points, to perform a given task on the point cloud (i.e., classification, retrieval or reconstruction). It is kept fixed during training and testing of \SNETwo. This ensures that sampling is being optimized to the task, rather than the task being optimized to an arbitrary sampling.

At the training phase, the generated points are fed to the task network. The points are optimized to the task at hand by minimizing the task loss. We use an additional sampling regularization loss term, that encourages each of the generated points to be close to one of the input points and forces the generated points to spread over the input cloud.

At inference, the generated points are matched with the input point cloud in order to obtain a subset of it. These are the sampled points, the final output of our process. These points are passed through the task network and its performance is evaluated. 

We present two sampling versions: \SNET and ProgressiveNet. In the first version (Figure~\ref{fig:overall_method}), we train a different sampling network per sample size. In the second one (Figure~\ref{fig:progressive_training}), we train one network that can be used to produce any sample size smaller than the input size.

\subsection{\SNET} \label{SampleNet_method}
The architecture of \SNET follows that of Qi \etal~\cite{qi2017pointnet}. The input points undergo a set of $1 \times 1$ convolution layers, resulting in a per point feature vector. Then, a symmetric feature-wise max pooling operation is used to obtain a global feature vector. Finally, we use several fully-connected layers. The output of the last layer is the set of generated points.

Let us denote the generated point set as $G$ and the input point set as $P$. We construct a sampling regularization loss, composed out of three terms:

\begin{equation}\label{eq:reg_loss_f}
L_f(G,P)=\frac{1}{|G|}\sum_{g \in G}{\min_{p \in P}||g-p||_2^2}
\end{equation}

\begin{equation}\label{eq:reg_loss_m}
L_m(G,P)=\max_{g \in G}{\min_{p \in P}||g-p||_2^2} \\
\end{equation}

\begin{equation}\label{eq:reg_loss_b}
L_b(G,P)=\frac{1}{|P|}\sum_{p \in P}{\min_{g \in G}||p-g||_2^2}.
\end{equation}


\noindent $L_f$ and $L_m$ keeps the points in $G$ close to those in $P$, in the average and worst case, respectively. This is designed to encourage tight matches in the following matching process. We found that mixing average and maximum operations speeds up convergence. $L_b$ ensures that the generated points are well spread over the input points, decreasing the number of collisions in the matching process. The sampling regularization loss is a weighted sum of these three terms:
\begin{equation} \label{eq:sampling_loss}
\begin{split}
L_s(G,P)=L_f(G,P) + \beta L_m(G,P)& \\
         + (\gamma + \delta|G|) L_b(G,P)&.
\end{split}
\end{equation}

\noindent Note that this is a generalization of the Chamfer distance~\cite{fan2017point_set_generation}, achieved when $\beta = 0$, $\gamma = 1$ and $\delta = 0$.

In addition, we denote $L_{task}$ as the task network loss. The total \SNET loss is:
\begin{equation} \label{eq:SampleNet_loss}
L^{\SNET}(G,P)= L_{task}(G) + \alpha L_s(G,P)
\end{equation} where $\alpha$ controls the regularization trade-off. The output of \SNET is a $k \times 3$ matrix, where $k$ is the sample size, and we train separately for each $k$.

\subsection{Matching} \label{matching_method}
The generated points $G$ are not guaranteed to be a subset of the input points $P$. In order to get a subset of the input points, we match the generated points to the input point cloud.

A widely used approach for matching two point sets is the Earth Mover's Distance (EMD)~\cite{achlioptas2018latent_pc, yuan2018pcn, zhang2018upsampling_pc, fan2017point_set_generation}. EMD finds a bijection between the sets that minimizes the average distance of corresponding points, while the point sets are required to have the same size. In our case, however, $G$ and $P$ are of different size.

We examine two matching methods. The first adapts EMD to uneven point sets. The second is based on nearest neighbour (NN) matching. Here we describe the latter, which yielded better results. The reader is referred to the supplementary material for details about the other matching method.

In NN-based matching, each point $x \in G$ is replaced with its closest euclidean corresponding point $y^* \in P$:
\begin{equation} \label{eq:nn_matching}
y^*=\argmin_{y \in P} {||x-y||_2}.
\end{equation}
\noindent Since several points in $G$ might be closest to the same point in $P$, the number of unique sampled points might be smaller than the requested sample size. Therefore, we remove duplicate points and get an initial sampled set. Then, we complete this set, up to the size of $G$, by running farthest point sampling (FPS)~\cite{qi2017pointnetplusplus}, where in each step we add a point from $P$ that is farthest from the current sampled set.

The matching process is only applied at inference time, as the final step of inference. During training, the generated points are processed by the task network as-is, since the matching is not differentiable and cannot propagate the task loss back to \SNET.
\subsection{ProgressiveNet: sampling as ordering}
\SNET is trained to sample the points to a single, predefined, sample size. If more than one sample size is required, more than one \SNET needs to be trained. But what if we want to train one network that can produce any sample size, i.e., sample the input at any sampling ratio? To this end we present ProgressiveNet. ProgressiveNet is trained to take a point cloud of a given size and return a point cloud of the same size, consisting of the same points. But, while the points of the input are arbitrarily ordered, the points of the output are ordered by their relevance to the task. This allows sampling to any sample size: to get a sample of size $k$, we simply take the first $k$ points of the output point cloud of ProgressiveNet and discard the rest. The architecture of ProgressiveNet is the same as \SNET, with the last fully connected layer size equal to the input point cloud size (has $3n$ elements).

\begin{figure}[h]
\begin{center}
\includegraphics[width=1.0\linewidth]{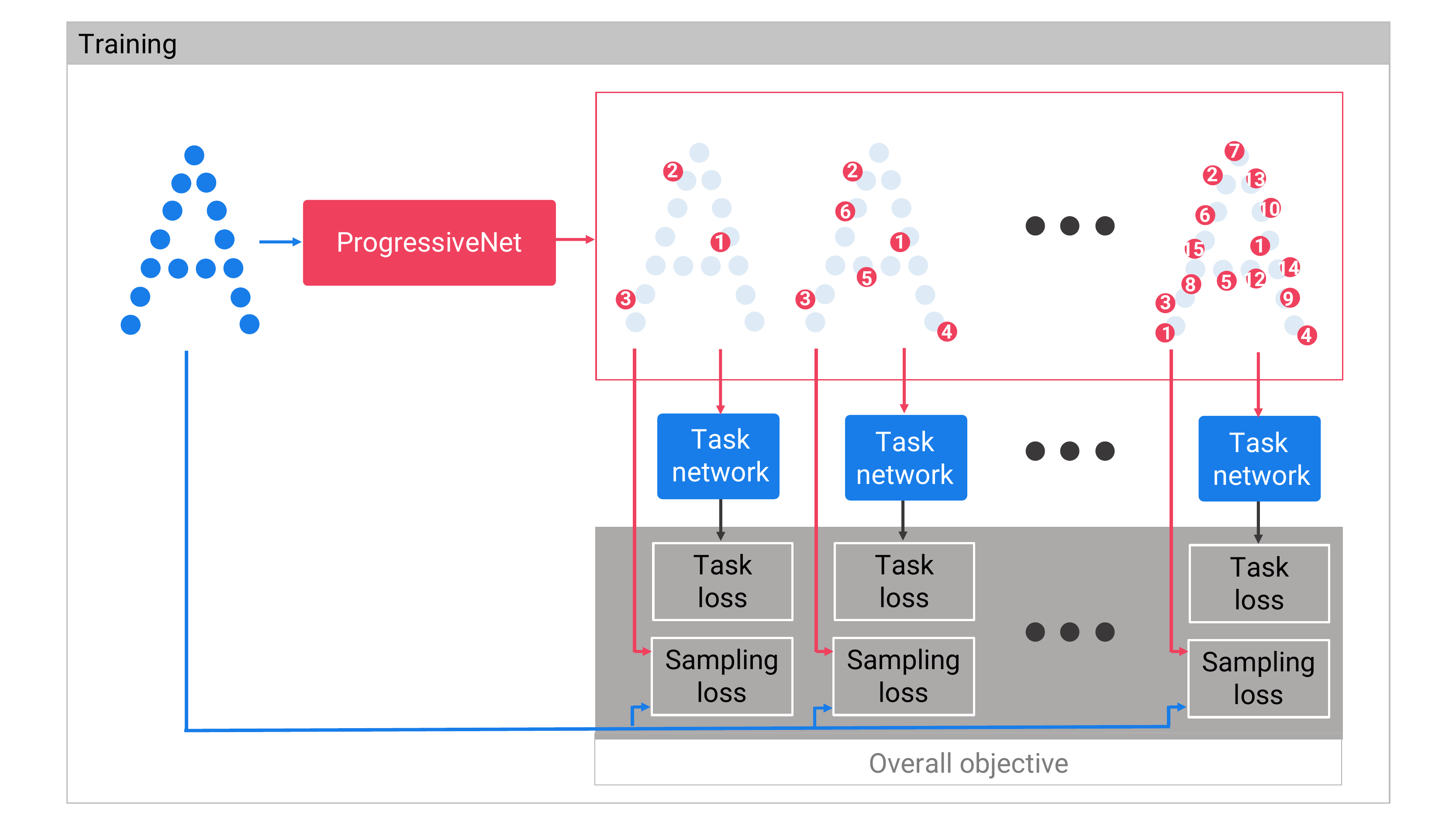}
\end{center}
\caption{{\bfseries ProgressiveNet training.} The generated points are divided into groups of increasing size, such that each group is a subset of the following larger group. Each group has corresponding task and sampling losses. The overall objective is the sum of the per-group losses. The task network was pre-trained and is kept fixed.}
\label{fig:progressive_training}
\end{figure}

To train ProgressiveNet (Figure~\ref{fig:progressive_training}) we define a set of sizes $C_s = \{2^1, 2^2, ... , 2^{log_{2}(n)}\}$. For each size $c \in C_s$ we compute a task loss term and a sampling regularization loss term, such that the total ProgressiveNet's loss becomes:  
\begin{equation}
L^{ProgressiveNet}(G,P)= \sum_{c \in C_s}{L^{\SNET}(G_c,P))}
\end{equation} where $L^{\SNET}$ is the loss term as defined in equation \ref{eq:SampleNet_loss} and $G_c$ are the first $c$ points from the points generated by ProgressiveNet (the first $3c$ elements of the output layer). 
This loss function defines a nested structure of point subsets. For example, the first subset, consisting of two points, is used in all terms $L^{\SNET}(G_c,P)$, for $c \geq 2$. Because this subset is used in all terms, it is contained in all larger subsets. Similarly, the first $4$ points define the second subset, that includes the first subset and is part of all larger subsets. 

Under this training process, the first $k$ generated points (for any $k$) are optimized to be suitable both for the task at hand as a stand-alone set and for integration with their preceding points to create a larger set that will improve results on the given task. This makes sure that a point that is more important for the task will appear earlier in the generated point set, while extra points will give diminishing marginal utility, resulting in a a task-oriented progressive decomposition of the point cloud~\cite{singh2006progressive_decomposition}.

At inference, the generated points (ProgressiveNet's output layer) are matched with the input point cloud using the same matching process we used in Section \ref{matching_method} for \SNETwo. We transfer the order of the generated points to their matched points. To obtain a specific sample size $k$, we take the first $k$ sampled points.


\section{Results}
We applied \SNET to three tasks: point set classification~\cite{qi2017pointnet}, retrieval and reconstruction~\cite{achlioptas2018latent_pc}. For classification and retrieval we adopt the ModelNet40~\cite{modelnet} point cloud data provided by Qi \etal~\cite{qi2017pointnet} and PointNet~\cite{qi2017pointnet} as the task network. For reconstruction we employ ShapeNet Core55 repository~\cite{shapenet} and the point set autoencoder of Achlioptas~\etal~\cite{achlioptas2018latent_pc}. 

Random sampling and FPS are employed as alternative non-data driven sampling approaches for comparison with our suggested approach. In order to make a fair comparison, we only use S-NET points after the matching process, so both our sampled points and the alternative approaches' points are subsets of the input point cloud. Further experimental details can be found in the supplementary material.

\subsection{Classification} \label{sec:classification_results}
We report instance classification results on the ModelNet40 data set, adopting the official train-test split. We employed the full version of PointNet as the task network for \SNET and the vanilla version of PointNet for ProgressiveNet (to save training time), except where otherwise noted.

\begin{figure}
\includegraphics[width=\columnwidth]{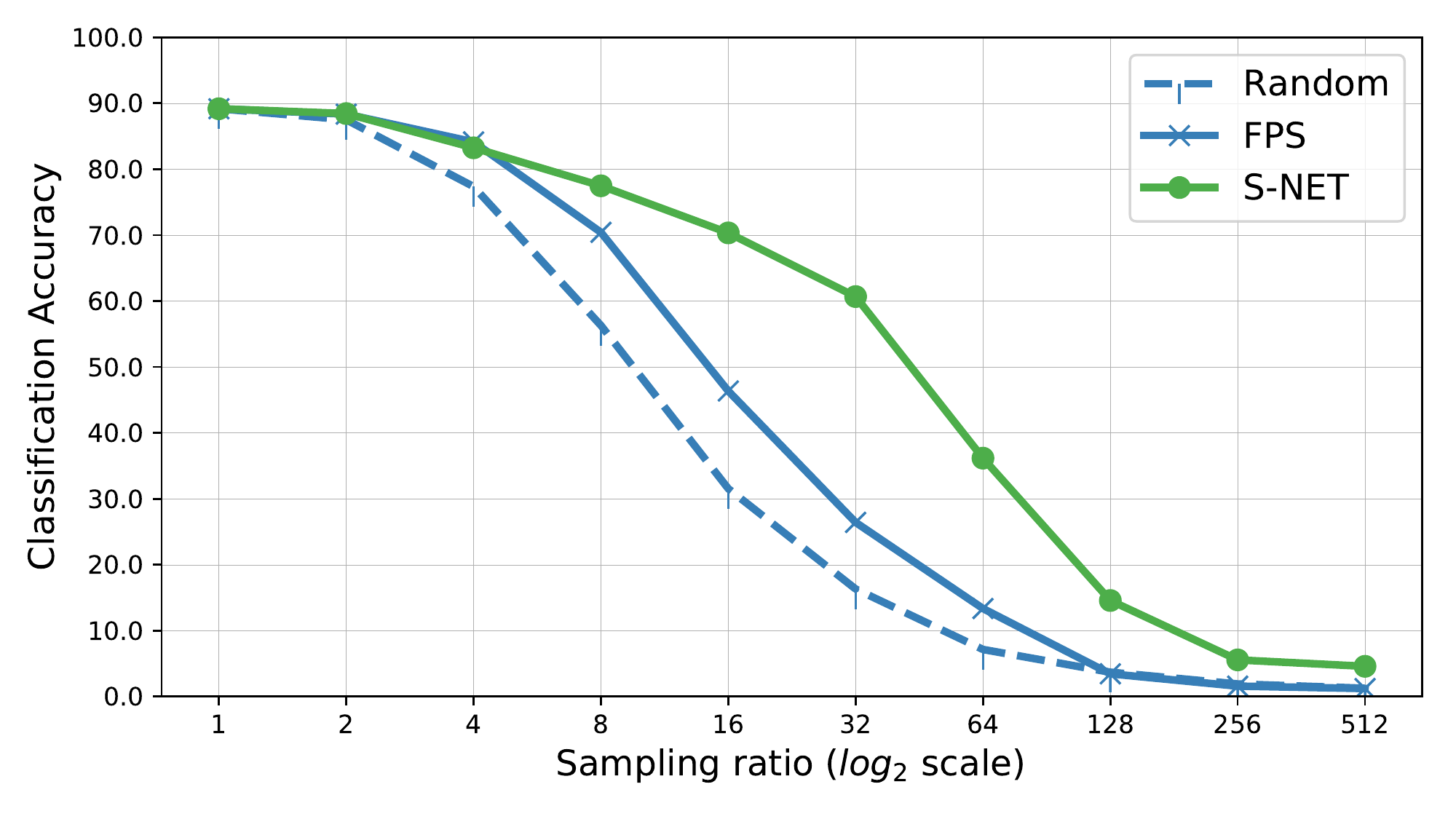}
\caption{{\bfseries \SNET for classification.} PointNet was trained on complete point clouds (1024 points) and evaluated on sampled point clouds of the test set using different sampling methods: random, FPS, and \SNETwo. The accuracy using \SNET is evidently higher.}
\label{fig:SampleNet_results}
\end{figure}

\medskip
\noindent {\bfseries \SNET} \quad Figure~\ref{fig:SampleNet_results} shows the classification accuracy of PointNet, when trained on the complete data (1024 points per point cloud) and tested on samples of different size. We compare several different sampling methods: random sampling with uniform distribution over the input points, FPS and \SNETwo. We trained 10 different \SNETwo s, for sample sizes of $k \in \{2,4,8,\dots,1024\}$ points. The sampling ratio is defined as $n/k = 1024/k$. We observe that the classification accuracy using \SNETwo's sampled points is equal or better than that of using FPS's points for any sampling ratio, with a margin of up to 34.2\% (for sampling ratio 32). 

Table~\ref{table:ReTrain} shows that the accuracy of PointNet is also higher when {\em training} it on the points sampled by \SNETwo. Each of the numbers in this table represents a different PointNet, each trained and tested on point clouds of a specific size that was sampled with a different sampling method. We see, for example, that PointNet that was both trained and tested on point clouds of just 16 points, achieved 85.6\% accuracy, when using \SNETwo, compared to only 76.7\% with FPS. This shows that \SNET is not over-fitted to the specific instance of the classifier it was trained with. Instead, it samples the points in a way that makes the sampled set easy to classify, creating a strong distinction between different classes in the data.  

\begin{figure}
\includegraphics[width=\columnwidth]{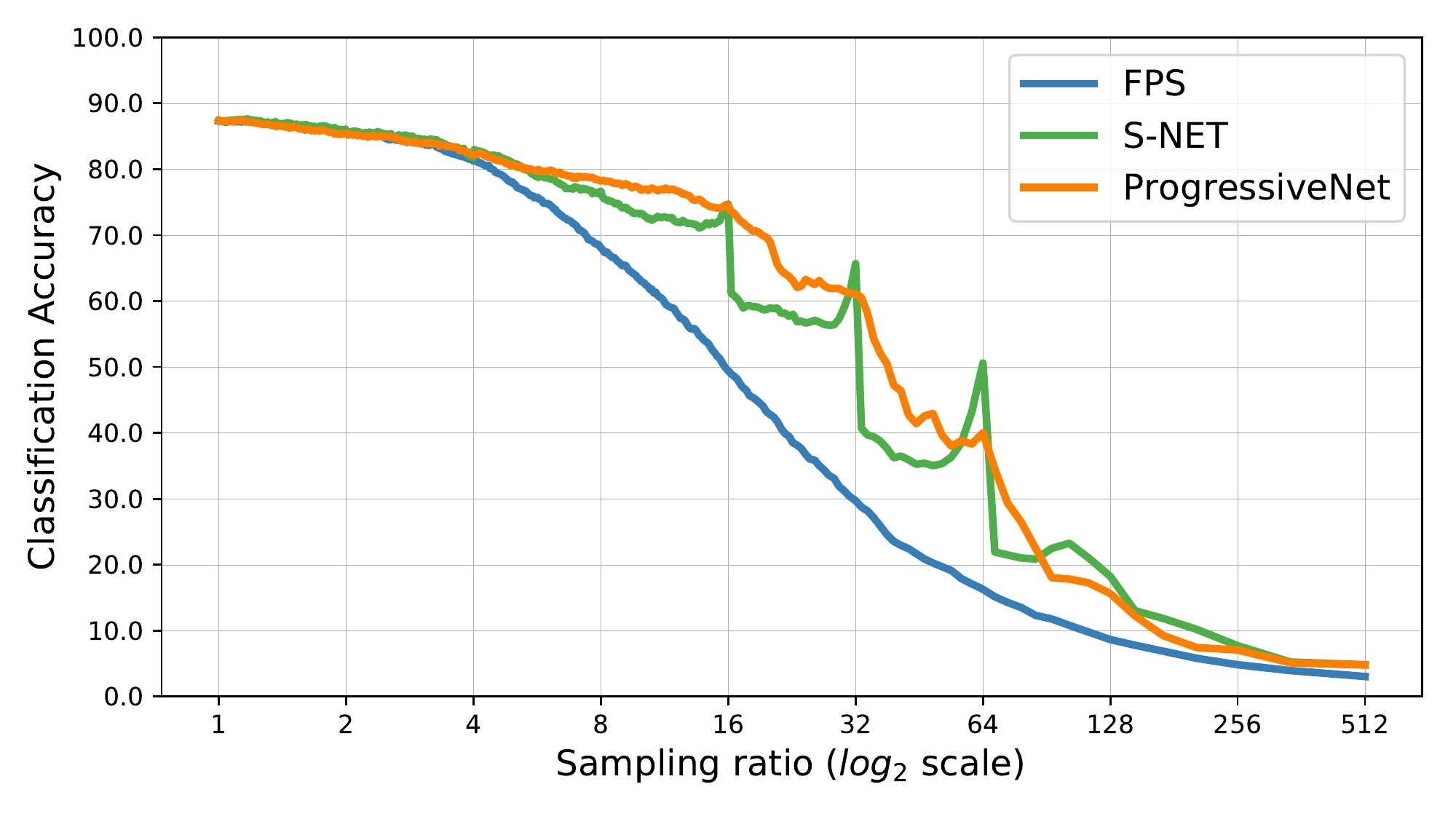}
\caption{{\bfseries ProgressiveNet vs. \SNETwo.} PointNet vanilla was trained on complete point clouds (1024 points) and evaluated on every sample size $k$ in the range $[2,1024]$. The sampling ratio is defined as $1024/k$. We compared three different sampling methods: FPS, \SNET and ProgressiveNet. \SNET is better for the sample sizes it was trained for, while ProgressiveNet performs better for sample sizes in between.}
\label{fig:ProgressiveNet_results}
\end{figure}

\begin{table}
\begin{center}
\begin{tabular}{ l c c c }
\hline
\#Sampled points & Random & FPS & \SNET \\
\hline\hline
1024 & 89.2 & 89.2  & 89.2 \\
512 & 88.2 & \bfseries 88.3 & 87.8 \\
256 & 86.6 & 88.1 &  \bfseries 88.3 \\
128 & 86.2 & 87.9 &  \bfseries 88.6 \\
64 & 81.5 & 86.1  & \bfseries 87.7 \\
32 & 77.0 & 82.2 & \bfseries 87.3 \\
16 & 65.8 & 76.7 & \bfseries 85.6 \\
8 & 45.8 & 61.6 & \bfseries 83.6 \\
4 & 26.9 & 35.2  & \bfseries 73.4 \\
2 & 16.6 & 18.3  & \bfseries 53.0 \\
\hline
\end{tabular}
\end{center}
\caption{{\bfseries Training PointNet classifier on sampled point clouds.} We sample the train and test data with different sampling methods: random sampling, FPS, and \SNETwo. Afterwards, PointNet is trained and evaluated on the sampled data. Applying \SNET allows good classification even with minimal data to train on. We conclude that \SNET transforms the data into a more separable representation.}
\label{table:ReTrain}
\end{table}
\medskip
\noindent {\bfseries ProgressiveNet}\quad In Figure~\ref{fig:ProgressiveNet_results} we compare the accuracy of PointNet vanilla on points sampled by \SNET (trained with PointNet vanilla, in this case) to those sampled by ProgressiveNet. The evaluation was done for all sample sizes in the range $[2, 1024]$. \SNET results are from 10 different \SNETwo s, each trained for a specific sample size, for sizes of $k \in \{2,4,8,\dots,1024\}$ points. For the sample sizes in between those values, we took the \SNET that was trained for a lower sample size and then completed with FPS to the required size. For example, to get a sample of size 48, we took the points sampled by \SNET that was trained to sample 32 points, and then made 16 steps of FPS. The Progressive results are from one ProgressiveNet with output size 1024, that was trained with classification and sampling loss terms for sizes $C_s = \{2, 4, 8, \dots , 1024\}$.

We observe that \SNET has better performance for the sample sizes it was trained for, while ProgressiveNet performs better for any sample size in between. This shows the advantage of ProgressiveNet, which orders the points by priority, so that the accuracy is approximately monotonically increasing in the sample size.

\begin{figure}
\includegraphics[width=\columnwidth]{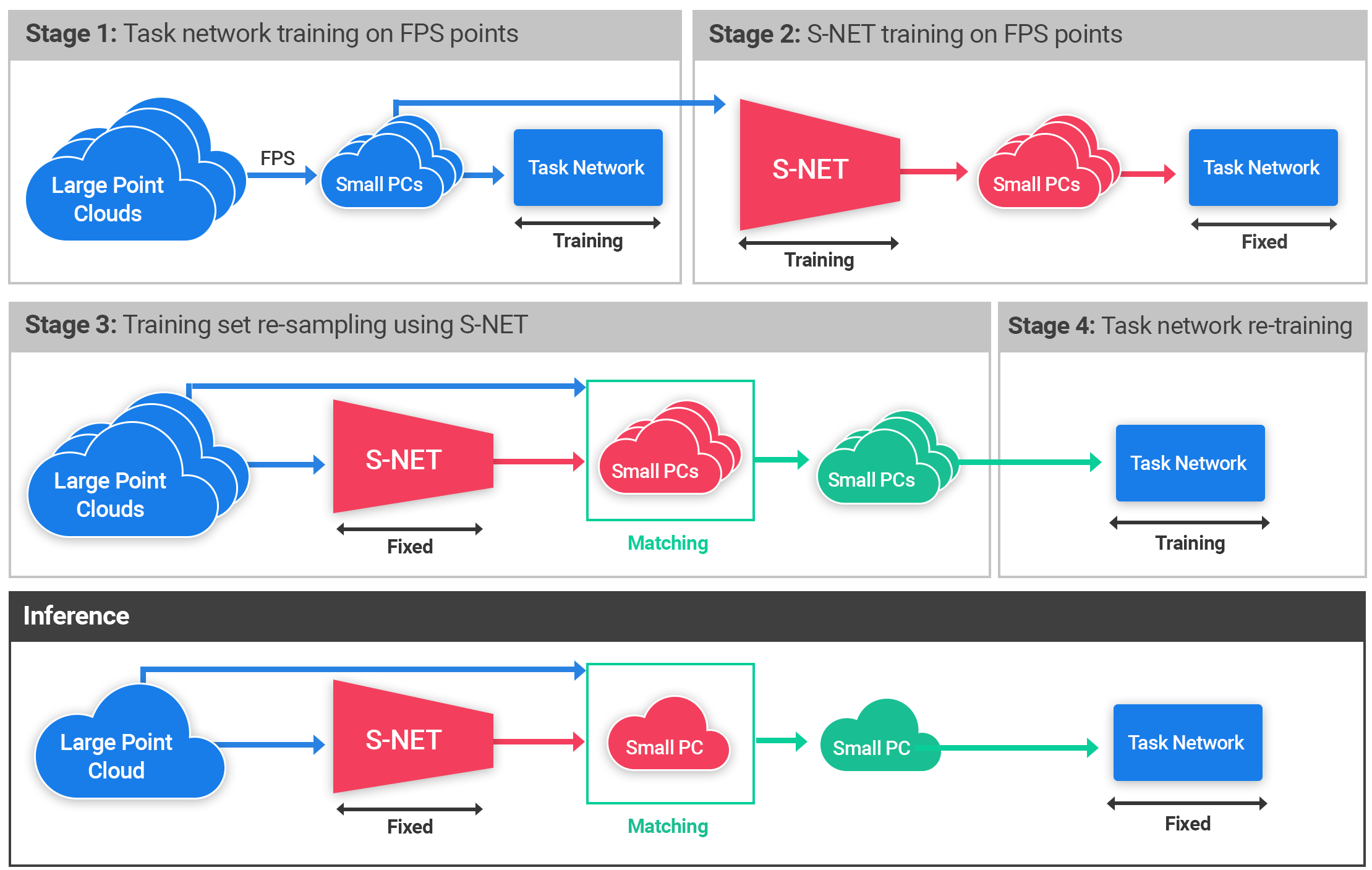}
\caption{{\bfseries Scalability by \SNETwo.} Illustration of the proposed training and inference procedure on large point clouds. In stage 1 we use FPS to sample the point clouds to a trainable size and use the sampled point clouds to train the task network. In stage 2 we train \SNET on the sampled point clouds, employing the fixed task network. In stage 3 we apply \SNET to re-sample the large point clouds and in stage 4 we train the task network again, this time on \SNETwo's points. At inference time, we use \SNET to sample a large point cloud to the size the task network was trained for.}
\label{fig:Large_PC_illustration}
\end{figure}

\medskip
\noindent {\bfseries Scalability by \SNET }\quad \SNET assumes that the task network is already trained. This will cause a problem in case of very large point clouds. We show that it is possible to train the task network on FPS-sampled point clouds and use that to train \SNET. The resulting \SNET can then be used to re-train the task network to a higher accuracy. Please see Figure~\ref{fig:Large_PC_illustration} for an illustration of the proposed training and testing procedure.

The following small scale simulation demonstrate this: We used FPS to sample ModelNet40 to $k=32$ points per point cloud, and trained PointNet on those sampled points. This is our baseline. The accuracy of the baseline on the test set is 82.2\%. When we fed this trained PointNet with point clouds of size $1024$, the accuracy was just 46.6\%. We then trained \SNET, using the $32$-sized point clouds training set, and the baseline PointNet as the task network. Then we sampled ModleNet40 again to size $k=32$, this time using \SNETwo. Finally, we trained PointNet on the points sampled by \SNETwo. The accuracy of the re-trained PointNet on the test set improved to 86.0\%. Employing \SNET allows us to improve PointNet's accuracy without training on larger point clouds. 

\medskip
\noindent {\bfseries Time and space considerations}\quad The time complexity of PointNet-like networks is dominated by their per-point convolution layers, and thus is strongly dependent on the size of the input point cloud. The space complexity of S-NET is a linear function of its output size $k$. \SNET offers a trade-off between space (number of parameters) and inference time (number of floating point operations). For example, cascading \SNET that samples a point cloud of 1024 to 16 points with a following PointNet reduces inference time by over 90\% compared to running PointNet on the complete point cloud, with only 5\% increase in space. See full details in the supplementary material. 

\begin{figure}
\includegraphics[width=\columnwidth]{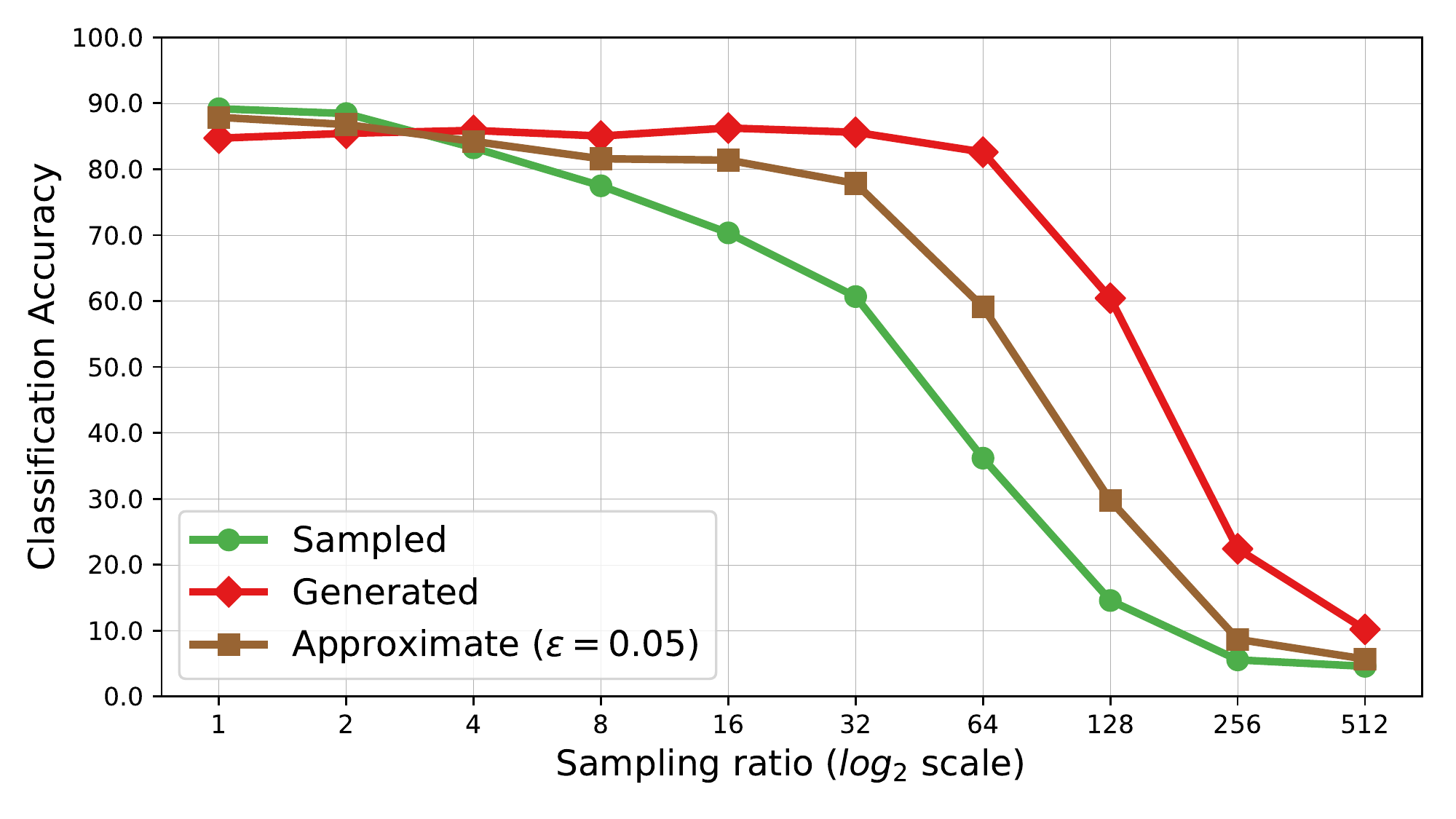}
\caption{{\bfseries Approximate sampling.} PointNet was trained on complete point clouds (1024 points) and evaluated on point clouds of different sizes. We use \SNETwo's generated and sampled points, as well as a third set of points, where each point is an interpolation between the generated and sampled point, bounded to be no more than $\epsilon = 0.05$ away from an original input point. Approximate sampling enables higher accuracy, when deviating from original points is not a concern.}
\label{fig:approximate_sampling}
\end{figure}

\medskip
\noindent {\bfseries Approximate sampling}\quad Up to this point we applied the matching post-processing step to compare ourselves directly with FPS. However, there might be settings where the $k$ output points do not have to be a subset of the original point cloud. In such cases, we can use \SNETwo's generated points directly, forgoing the matching step. One can use either the generated or the sampled points. A third alternative is to interpolate between the two, i.e., using points that are up to $\epsilon$ away from the original input points. This is done by interpolating each generated point with its matched sampled point, to get a third point on the line between them, no more than $\epsilon$ away from the sampled point. Figure~\ref{fig:approximate_sampling} shows that PointNet's accuracy is higher when feeding it the generated points. For point clouds normalized to the unit sphere, we find that choosing $\epsilon=0.05$ results in classification accuracy that is about mid-way between the accuracy on the sampled and generated points. Note that $\epsilon$ is set at inference time.

\medskip
\noindent {\bfseries Critical set sampling}\quad The critical set, as defined by Qi \etal~\cite{qi2017pointnet}, is the set of points that contributed to the max pooled features. A plausible alternative to our method might be to sample the critical points that contribute the most features. We tried this approach and found it viable only at small sampling ratios. See full details in the supplementary material.
\subsection{Retrieval} \label{sec:retrieval_results}
We now show that employing \SNET instead of FPS leads to better retrieval results. Specifically,  we took the \SNET that was trained with PointNet for classification as the task network, without re-training it. We sampled the points that are fed to PointNet and used its penultimate layer as a shape descriptor. Retrieval was done based on $L_2$ distance on this shape descriptor. We repeated the experiment with every shape in the ModelNet40 test set serving as a query shape. The experiment was repeated with different sample sizes for both \SNET and FPS.



\begin{figure}
\includegraphics[width=\columnwidth]{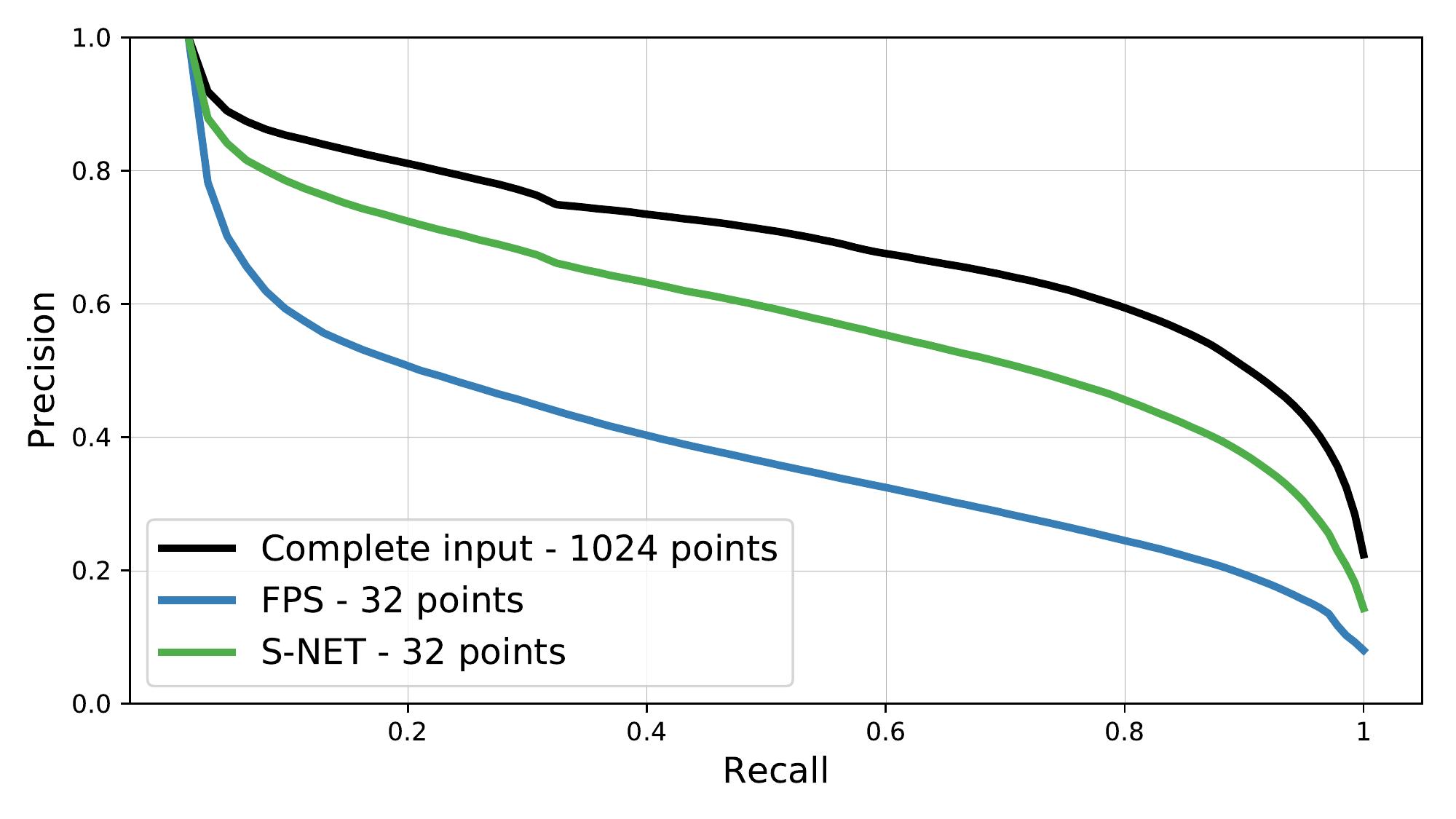}
\caption{{\bfseries Precision-Recall curve for point cloud retrieval.} We compare FPS and \SNET when sampling 32 points, as well as using the complete 1024 points data. Significantly better precision is achieved when using \SNET compared to FPS, across all recall values.}
\label{fig:Precision_Recall}
\end{figure}

Table~\ref{table:Precision_Recall} summarizes the mean Average Precision (mAP) for different sampling ratios and sampling methods. We see that \SNET performs much better for sampling ratios larger than 4 and is not very sensitive to the sampling ratio. Figure~\ref{fig:Precision_Recall} presents the precision-recall curve for the original data and for the sampled data using FPS and \SNETwo, sampling to $k$=32 points per shape. We observe significantly better precision when applying \SNET across all recall values.

\begin{table}
\begin{center}
\begin{tabular}{ l c c }
\hline
Sampling ratio & FPS mAP & \SNET mAP \\
\hline\hline
1   &	            71.3   &	            71.3 \\
2   &	\bfseries   70.1   &	            69.8 \\
4   &	\bfseries   65.7   &	            64.8 \\
8   &	            58.3   &	\bfseries   60.4 \\
16   &	            49.4   &	\bfseries   59.0 \\
32   &	            37.7   &	\bfseries   59.0 \\
64   &	            27.4   &	\bfseries   54.5 \\
\hline
\end{tabular}
\end{center}
\caption{{\bfseries Point cloud retrieval.} We took the same \SNET that was trained with PointNet classifier as the task network and applied it as the sampling method for retrieval. The shape descriptor was PointNet's activations of the layer before the last, with $L_2$ as the distance metric. We measured the macro mean Average Precision (mAP) for different sampling ratios and methods. \SNET performs better than FPS for large sampling ratios and is almost insensitive to the sampling ratio.}
\label{table:Precision_Recall}
\end{table}

\subsection{Reconstruction} \label{sec:reconstruction}
We next learn to sample with an autoencoder as the task network. We used the ShapeNet Core55 point cloud data provided by Achlioptas \etal~\cite{achlioptas2018latent_pc}, as well as their autoencoder and trained it on four shape classes: table, car, chair and airplane. These classes have the most available models. Each shape class is split into 85\%-5\%-10\% for train-validation-test sets. The autoencoder was trained to receive and reconstruct point clouds of 2048 points.


The reconstruction error of the autoencoder is measured by the Chamfer distance~\cite{achlioptas2018latent_pc}. In order to compare different sampling methods, we use Normalized Reconstruction Error (NRE). That is, we reconstruct the complete point set from a subset of points and from the complete point set and take the reconstruction error ratio between the two. We trained several \SNETwo s with the following sample sizes: $k \in \{16, 32, \dots, 2048\}$, and a single ProgressiveNet with loss terms for the same sizes. As an alternative sampling approach we used FPS.

\begin{figure}
\begin{center}
\includegraphics[width=1.0\linewidth]{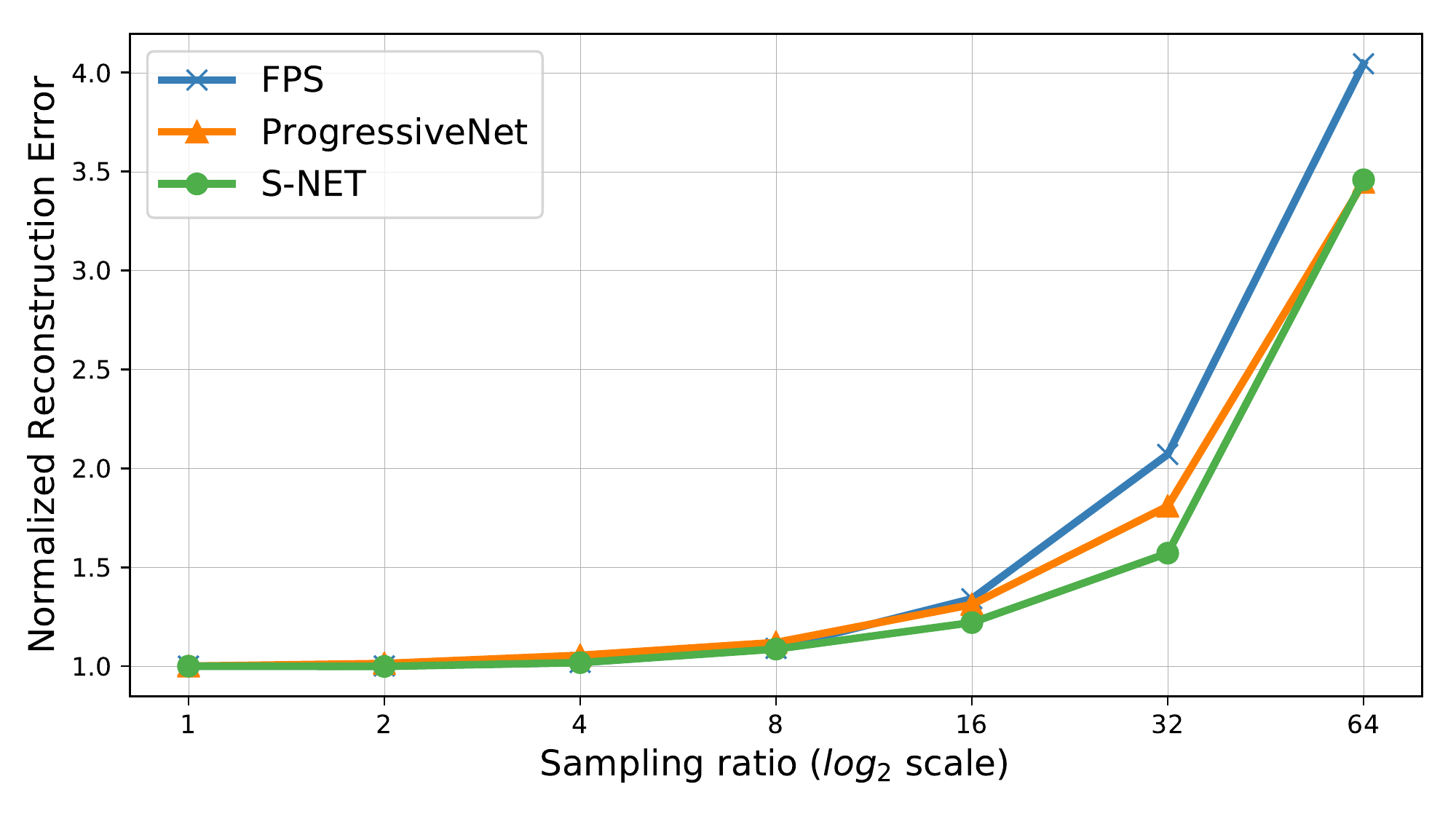}
\end{center}
\caption{{\bfseries Normalized Reconstruction Error (NRE).} We trained an autoencoder on complete point clouds of 2048 points, and evaluated the reconstruction error from sampled point clouds with FPS, \SNET and ProgressiveNet on the test split. Up to a sampling ratio of 8, the error for \SNET and ProgressiveNet is on par with FPS. However, at higher sampling ratios \SNET and ProgressiveNet achieves lower error.}
\label{fig:recon_loss_vs_samp_size_multi}
\end{figure}

\medskip
\noindent {\bfseries Normalized reconstruction error} \quad Figure~\ref{fig:recon_loss_vs_samp_size_multi} presents the NRE as a function of the sampling ratio, where the sampling ratio is defined as $n/k=2048/k$. We compare FPS with \SNET and ProgressiveNet. For small sampling ratios, the NRE for our sampled points is similar to that of FPS. However, as the sampling ratio increases, our sampling methods outperform FPS. For example, at sampling ratio of 32, the NRE of FPS is a little over $2$, while the NRE of \SNET and ProgressiveNet is about $1.5$ and $1.75$, respecitvely. \SNET achieves lower NRE than ProgressiveNet, since the former was optimized separately per sampling ratio, resulting in improved reconstruction performance. We learn to sample points from unseen shapes that enable lower reconstruction error.


\begin{figure}
\begin{center}
\begin{tabular}{ c c c }
\includegraphics[width=0.25\linewidth]{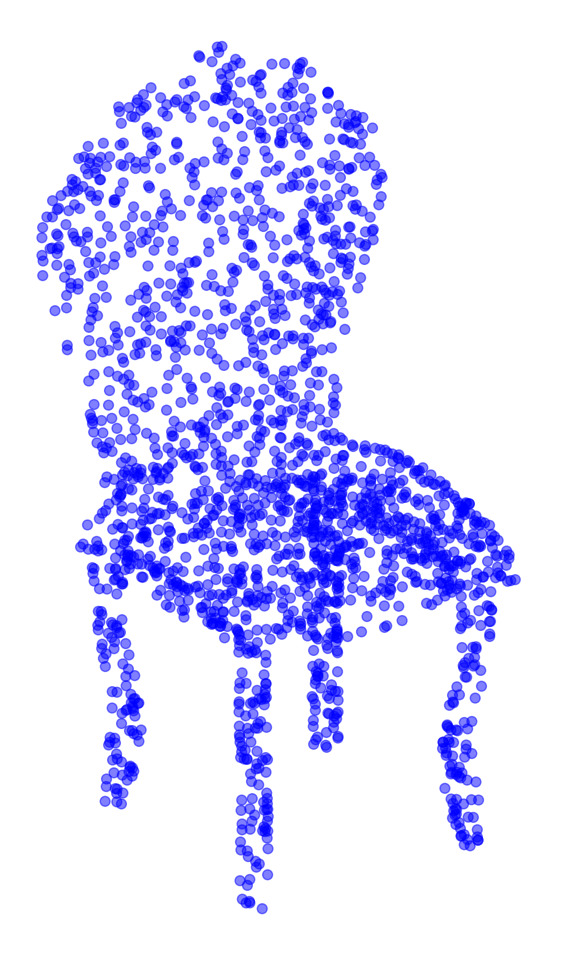} &
\includegraphics[width=0.25\linewidth]{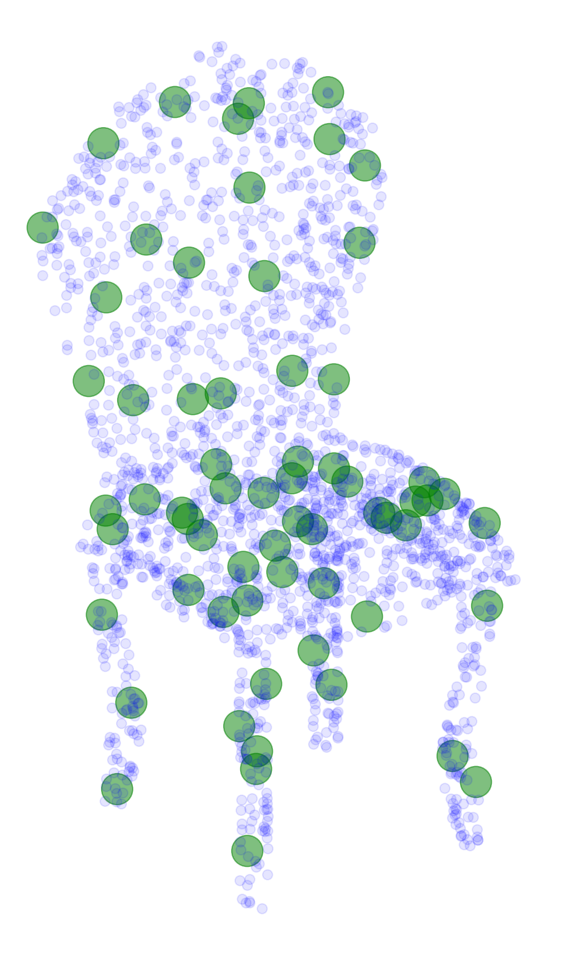} &
\includegraphics[width=0.25\linewidth]{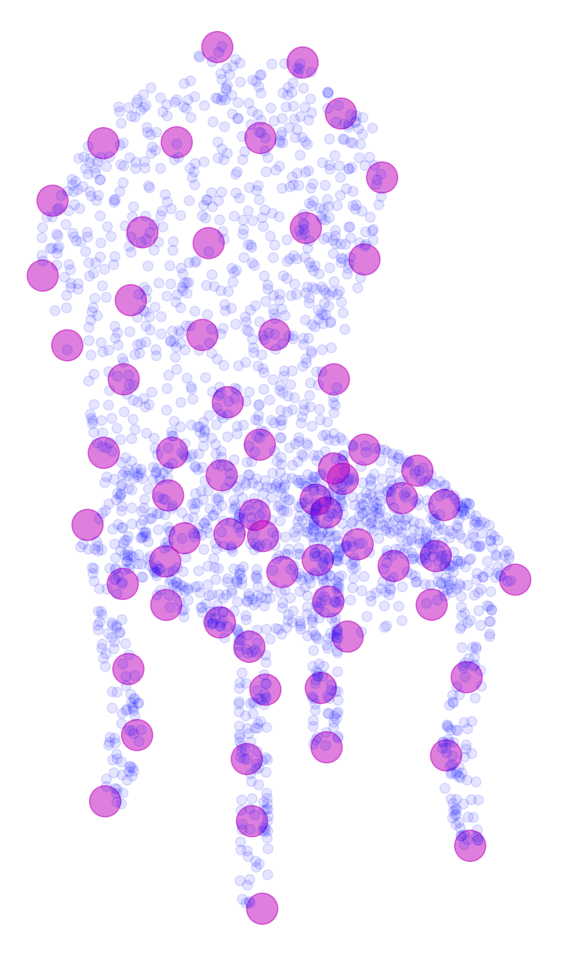} \\
\includegraphics[width=0.25\linewidth]{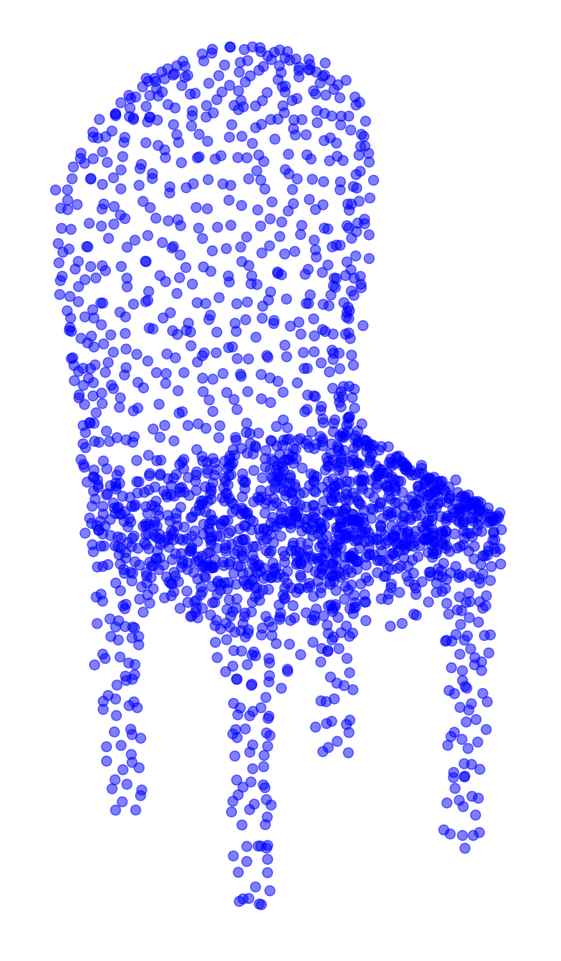} &
\includegraphics[width=0.25\linewidth]{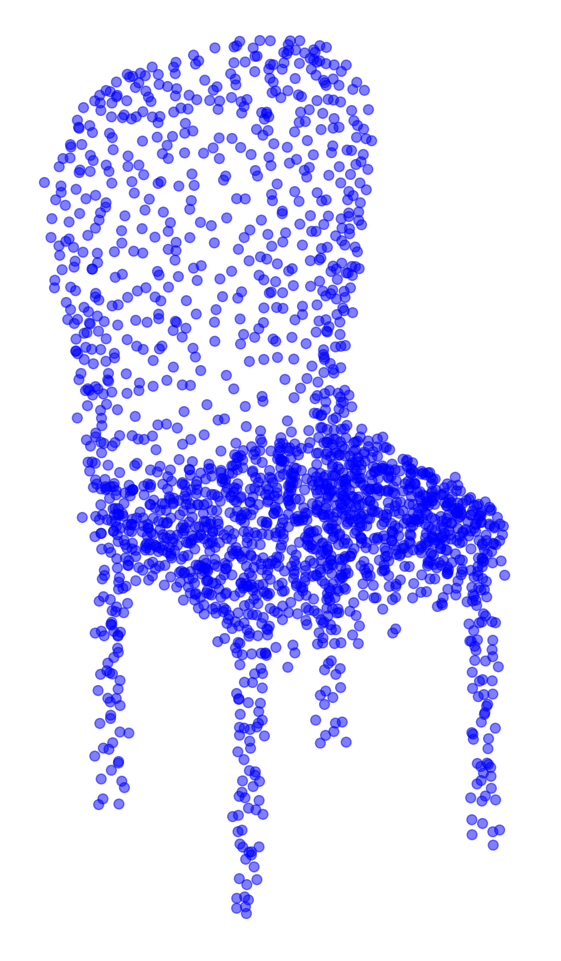} &
\includegraphics[width=0.25\linewidth]{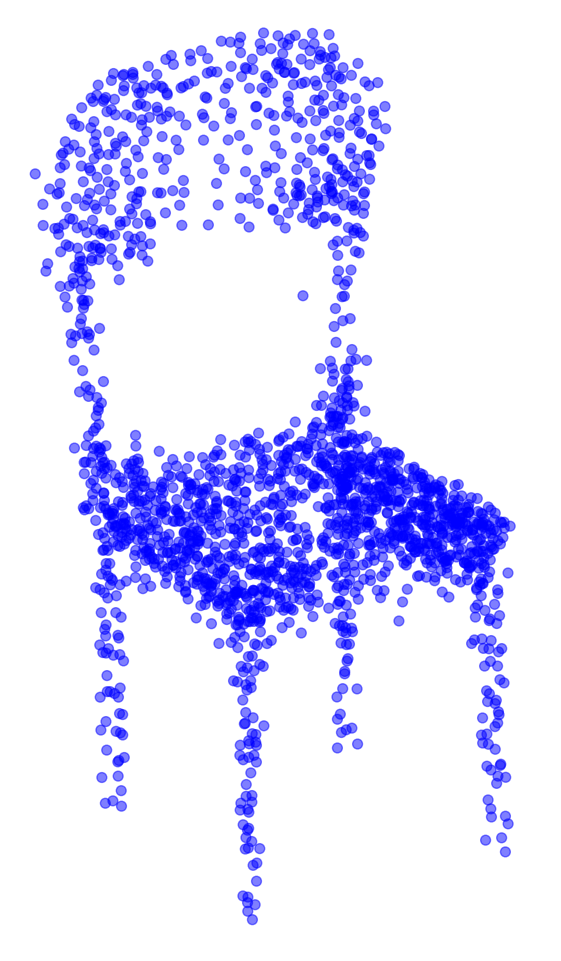} \\
Complete Input  &   \SNET & FPS \\
NRE = 1.00 &
NRE = 1.11 &
NRE = 2.38 \\
\end{tabular}
\caption{{\bfseries Point cloud reconstruction.} NRE stands for Normalized Reconstruction Error. Top row: complete input point cloud of 2048 points, input with 64 \SNET sampled points (in Green), input with 64 FPS points (in Magenta). The sampled and FPS points are enlarged for visualization purpose. Bottom row: reconstructed point cloud from the input and from the corresponding sample. The reconstructed point cloud from \SNETwo's sampled points is visually more similar to the input and has lower reconstruction error.}
\label{fig:pc_recon}
\end{center}
\end{figure}

\medskip
\noindent {\bfseries Sample and reconstruction visualization} \quad Figure~\ref{fig:pc_recon} compares the reconstruction result from the entire point cloud, to that from a sample size of 64 points, where the samples are produced by either \SNET or FPS. The reconstruction quality when employing \SNET is higher than that of using FPS and approaches that of using the entire point cloud. Interestingly, the points sampled by \SNET are {\em non-uniformly} distributed, as opposed to the more uniform distribution of the points sampled by FPS.




\begin{figure}
\begin{center}
\begin{tabular}{ c c }
\includegraphics[width=0.45\linewidth]{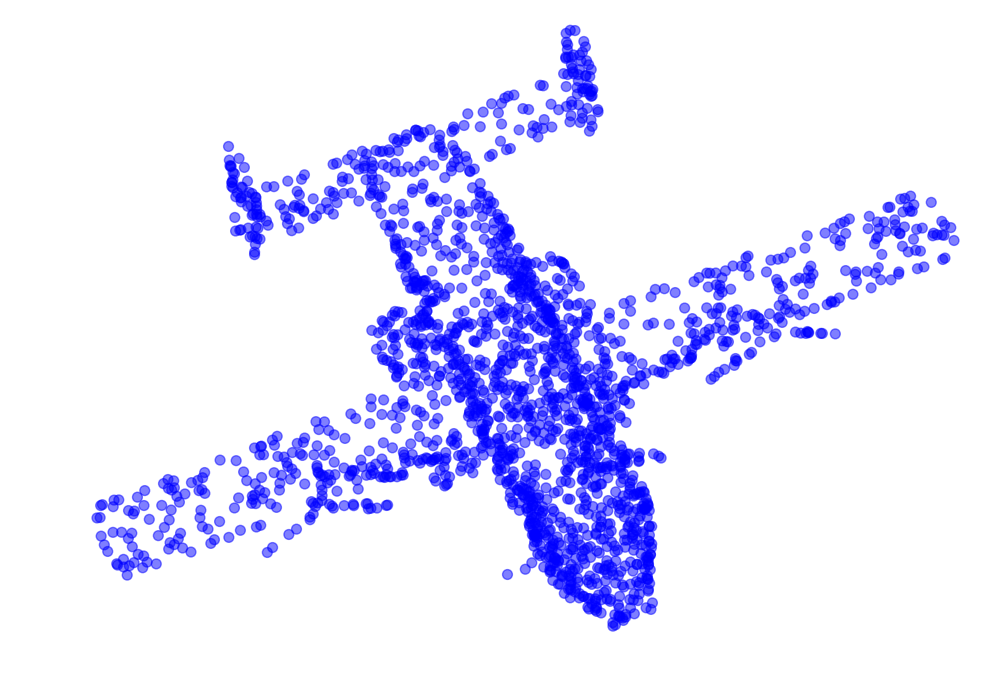} &
\includegraphics[width=0.45\linewidth]{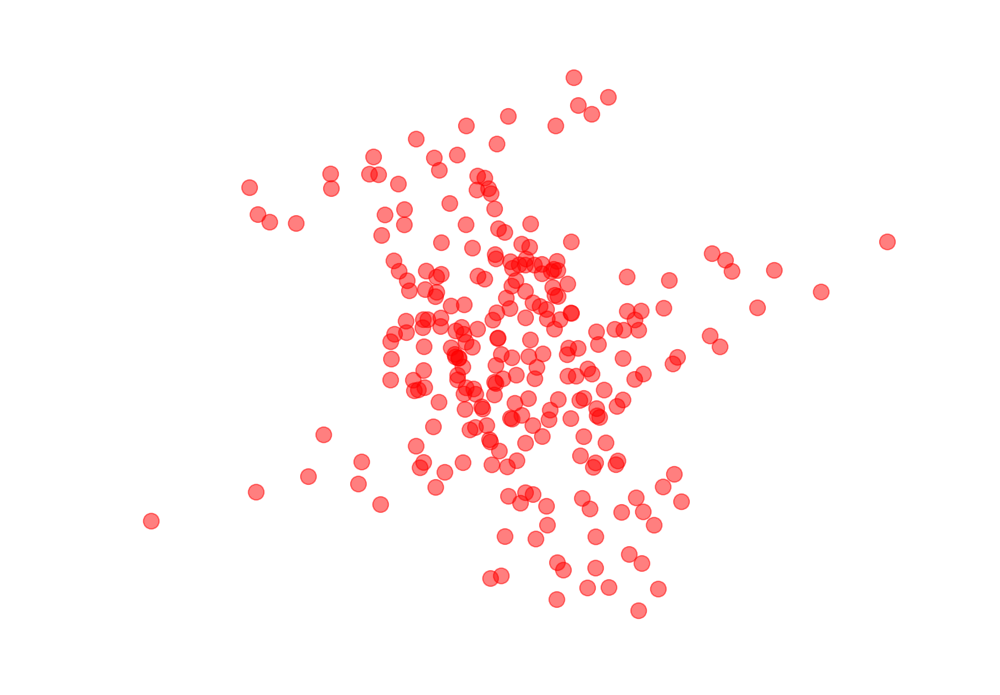} \\
\includegraphics[width=0.45\linewidth]{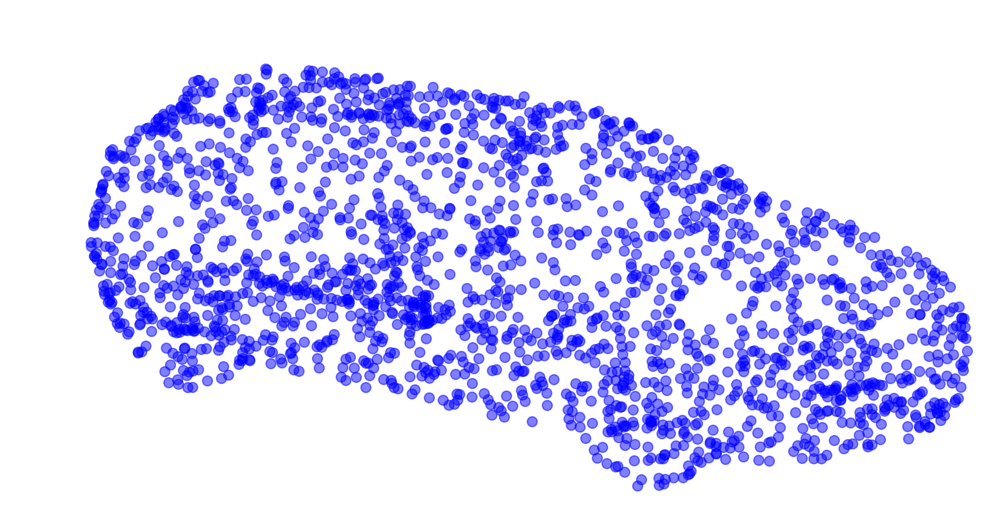} &
\includegraphics[width=0.45\linewidth]{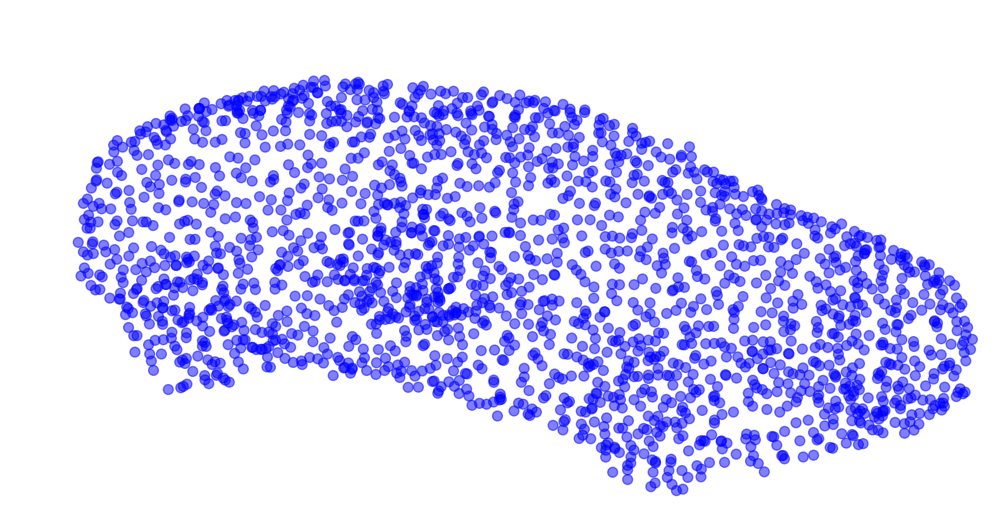} \\
\end{tabular}
\caption{{\bfseries Adversarial simplification.} Top row: input shape (in Blue) and 256 generated points (in Red). Bottom row: target shape and reconstruction from the generated points. While the simplified point cloud resembles the input airplane shape, it is reconstructed to a {\em completely different} shape - a car!}
\label{fig:adversarial_sampling}
\end{center}
\end{figure}

\medskip
\noindent {\bfseries Adversarial simplification} \quad In this proof of concept we show how to trick the autoencoder. We simplify a point cloud to be visually similar to one class but reconstructed by the autoencoder to a shape from a {\em different} class. We train \SNET with a single pair of input and target shapes, where the input is a shape from one class and the target is a shape from another class. The sampling loss was between the input and the points generated by \SNETwo. The reconstruction loss was between the target shape and the reconstructed one. Figure~\ref{fig:adversarial_sampling} shows the result of turning an airplane into a car.


\section{Conclusions}

We presented a method that learns how to sample a point cloud that is optimized for a downstream task. The method consists of a simplifying network, \SNETwo, followed by a post-processing matching step. We also suggested a network, termed ProgressiveNet, for ordering a point cloud according to the contribution of each point to the task. The resulting methods outperform FPS in sampling points for several tasks: classification, retrieval and reconstruction of point clouds.

The proposed method is general and can produce a small point cloud that consists of points that are not necessarily part of the original input shape. The output point cloud minimizes a geometric error with respect to the input point cloud while optimizing the objective function of a downstream task.
We have shown that learning to sample can improve results and be used in conjunction with various applications.




{\bf Acknowledgments:} Parts of this research were supported by ISF grant 1917/15.


{\small
\bibliographystyle{IEEEtran} 
\bibliography{references}
}

\newpage
\appendix
\section*{Supplementary}
In the following sections we provide additional details and analysis regarding our results.
In Section~\ref{sec:AdditionalResults} we provide experimental details and additional results for our method.
Section~\ref{sec:Examples} displays visual examples of retrieved shapes and progressive sampling. 
Section~\ref{sec:Extensions} offers a new application for the progressive concept: A progressive autoencoder.




\section{Additional results} \label{sec:AdditionalResults}
In this section we extend the results section of the paper to include: details of our experimental parameters, analysis of the different matching methods, evaluation of regularization parameters, elaboration of the space and time considerations, analysis of critical set sampling, and comparison of the class accuracy.

\subsection{Experimental details} \label{sec:ExperimentalDetails}

\noindent {\bfseries Classification Architecture}\quad The architecture of \SNET for the classification experiment is inspired by the vanilla version of PointNet by Qi \etal~\cite{qi2017pointnet}. We use per point $1 \times 1$ convolution layers with output sizes of $[64,\, 64,\, 64,\, 128,\, 128]$. Then, a feature-wise max-pooling layer is used to obtain a global feature vector. This feature vector is passed through four fully connected layers of size $[256,\, 256,\, 256,\, k \times 3]$, where $k$ is the sample size.. All convolution and fully connected layers are followed by ReLU non-linearity~\cite{nair2010relu} and batch-normalization layer~\cite{ioffe2015batch_normalization}, except for the output layer. ProgressiveNet takes the architecture of \SNET with $k=1024$.

\medskip
\noindent {\bfseries Classification Optimization}\quad We used Adam optimizer~\cite{kingma2015adam} with a learning rate of $0.01$, decay rate of $0.7$ every $60000$ steps and batch size of $32$ point clouds. The regularization weights (for equations~\ref{eq:sampling_loss} and~\ref{eq:SampleNet_loss}) were set to $\alpha=30,\, \beta=1,\, \gamma=1,\, \delta=0$ for \SNET and $\alpha=30,\, \beta=1,\, \gamma=0.5,\, \delta=1/30$ for ProgressiveNet. We trained the networks for 500 epochs with a GTX 1080 Ti GPU. When using PointNet as the task network, \SNET training takes between $2$ to $7$ hours, depending on the sample size. ProgressiveNet training takes $6$ hours, when using PointNet vanilla as the task network.

\medskip
\noindent {\bfseries Classification Augmentation}\quad We employed the augmentation proposed by Qi \etal~\cite{qi2017pointnet}: random rotation of the point cloud along the up-axis, and jittering the position of each point by a Gaussian noise with zero mean and $0.02$ standard deviation.


\medskip
\noindent {\bfseries Reconstruction Architecture}\quad The architecture of \SNET for the reconstruction experiment follows the same basic structure as in the classification case, with $1 \times 1$ of sizes $[64,\, 128,\, 128,\, 256,\, 128]$ and fully connected layers of seizes $[256,\, 256,\, k \times 3]$. ProgressiveNet takes the architecture of \SNET with $k=2048$.


\medskip
\noindent {\bfseries Reconstruction Optimization}\quad 
To train both \SNET and ProgressiveNet, we used Adam optimizer with a learning rate of $0.0005$, momentum $0.9$ and mini-batches of $50$ shapes. The regularization weights were set to $\alpha=0.01,\, \beta=1,\, \gamma=0,\, \delta=1/64$. For ProgressiveNet, the total loss is divided by the number of sample sizes $|C_s|$. We trained the networks for $500$ epochs with a Titan X Pascal GPU. \SNET training takes between $4$ to $8$ hours, depending on the sample size. ProgressiveNet training time is about $12$ hours. No augmentation was used for reconstruction.


\subsection{Matching methods} \label{sec:MatchingMethods}
We examined two matching methods. The first one is based on nearest neighbour (NN) matching, as detailed in Section~\ref{matching_method}. In the second one, we adapted the implementation of the $(1 + \epsilon)$ approximation scheme for EMD~\cite{bertsekas1985emd}, provided by Fan \etal~\cite{fan2017point_set_generation}. This implementation gives a matching score for each point in one point cloud to each point in the other one. We feed the algorithm with $G$ and $P$ and take the points from $P$ corresponding to the highest matching score for each point in $G$. In Table~\ref{table:Matching} we compare these matching methods. The accuracy is better with NN matching for large sampling ratios, and is almost the same for small ones.

NN matches every point in $G$ with its closest point in $P$. This ensures that all generated points that \SNET learned are reflected in the matched set. This is not the case when using $(1+\epsilon)$, where a generated point might be matched with a far input point, if it serves the $(1+\epsilon)$ approximation scheme. In addition, NN matching is independent for each point,  thus it guarantees to keep the order of the points produced by ProgressiveNet.

\begin{table}
\begin{center}
\begin{tabular}{ l c c c }
\hline
Sampling &  FPS & ProgressiveNet & ProgressiveNet \\
ratio & & $(1+\epsilon)$ matching & NN matching \\
\hline\hline
1               &	87.3    &	            87.3    &	            87.3    \\
2               &	85.6    &	\bfseries   85.7    &	            85.4    \\
4               &	81.2    &	\bfseries   82.4    &   	        82.3    \\
8               &	68.1    &	            76.4    &	\bfseries   78.2    \\
16              &	49.4    &	            68.3    &	\bfseries   74.4    \\
32              &	29.7    &	            50.7    &	\bfseries   61.0    \\
64              &	16.3    &	            27.8    &	\bfseries   40.0    \\
\hline
\end{tabular}
\end{center}
\caption{{\bfseries Matching methods comparison.} We  compare  the  classification accuracy of PointNet vanilla on the sampled points when using different matching methods: $(1+\epsilon)$ and Nearest Neighbour (NN). FPS is shown for reference. NN matching is better for all sampling ratios larger than 4.}
\label{table:Matching}
\end{table}

\subsection{Regularization} \label{sec:Regularization}
\begin{figure}
\includegraphics[width=\columnwidth]{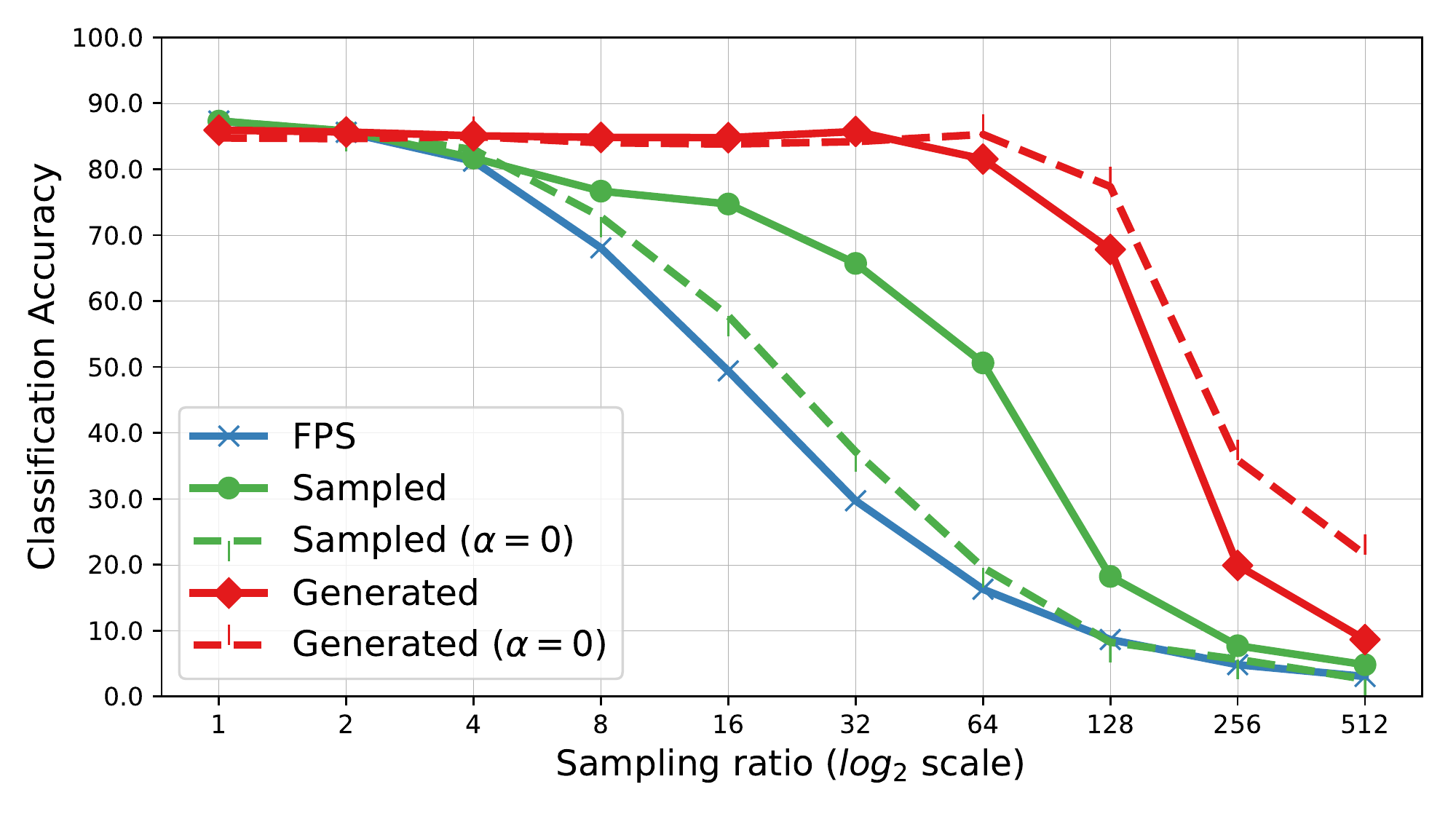}
\caption{{\bfseries Turning off the sampling regularization loss ($\alpha = 0$).} PointNet vanilla was trained on complete point clouds (1024 points) and evaluated on either \SNETwo's generated or sampled points, either with or without the sampling regularization loss. Without the regularization, the accuracy using the generated points is slightly increased. However, the accuracy when using the sampled points decreases substantially.}
\label{fig:zero_alpha}
\end{figure}

In this section we explore the influence of the sampling regularization loss and it's components. Please see Section~\ref{SampleNet_method} for the relevant background and equations.

\medskip
\noindent {\bfseries Turning off the regularization ($\alpha = 0$)}\quad 
Figure~\ref{fig:zero_alpha} shows the accuracy of PointNet vanilla using either \SNETwo's generated or sampled points, with and without regularization. Without the regularization, the accuracy using the generated points is slightly increased, which is expected, since the regularization sacrifices some of the classification loss to minimize the sampling loss. However, after the matching process the classification accuracy on the unregularized sampled points is much lower than when using the regularization. Without the sampling regularization, the generated points are not close to the original points and are not spread out over the whole shape. As a result, the matching process causes the sampled points to be very different from the generated points. Therefore, the sampled points do not preserve the contribution of the generated points to the task.

\begin{figure}
\includegraphics[width=\columnwidth]{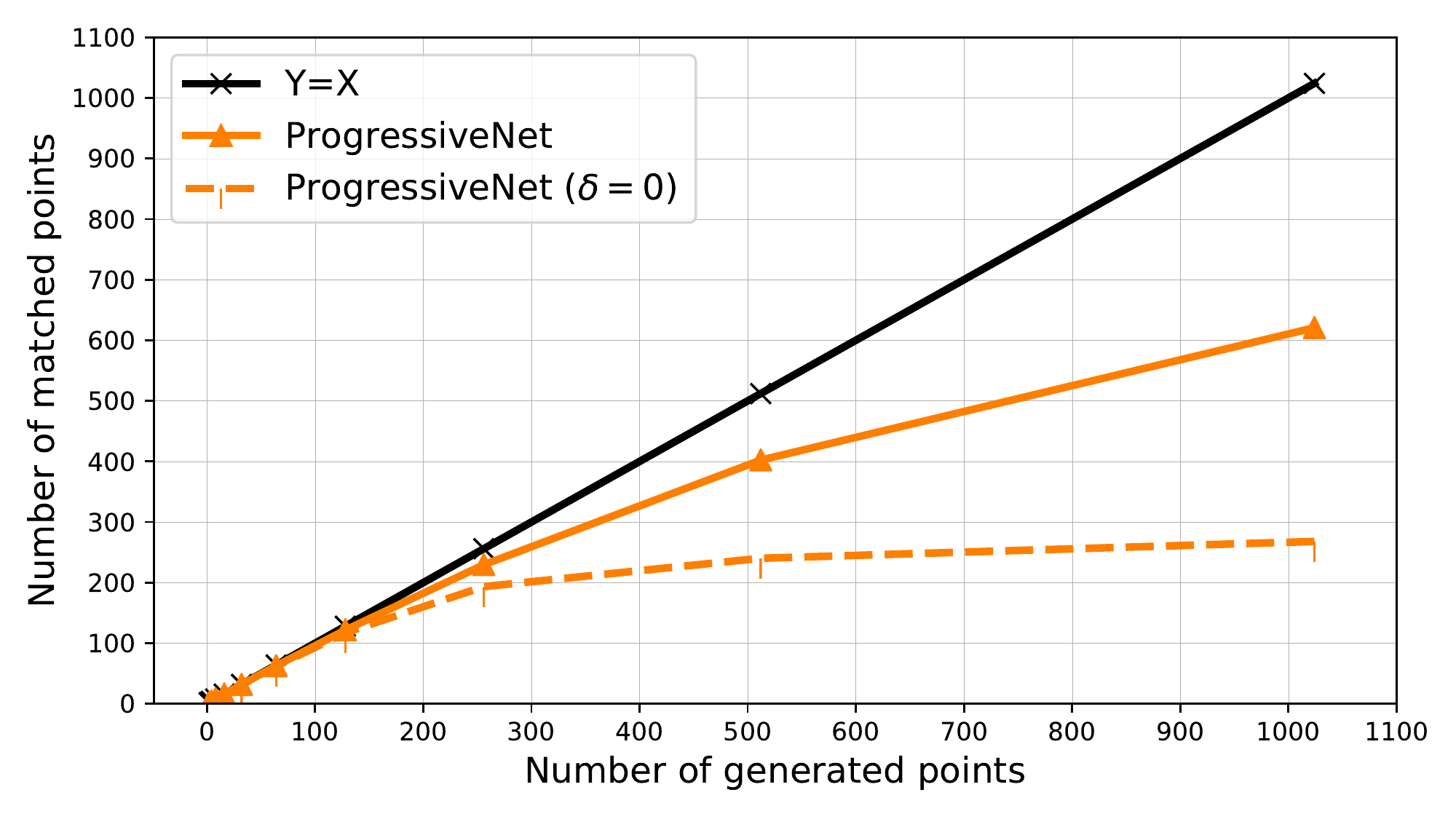}
\caption{{\bfseries Turning off the linear factor for $L_b$ component ($\delta = 0$).} Average number of unique points in the initial sampled set (after NN matching, before FPS completion) is shown as a function of the number of generated points. Ideally, we would like to get the line $Y=X$, which means that every generated point has a unique matched input point. Setting $\delta = 0$ greatly reduce the number of unique matches.}
\label{fig:zero_delta}
\end{figure}

\medskip
\noindent {\bfseries Turning off the linear factor for $L_b$ component ($\delta = 0$)}\quad In Figure ~\ref{fig:zero_delta} we see the average number of unique points in the initial sampled set (after NN matching, before FPS completion) as a function of the number of generated points. For low values, the numbers are almost the same, but they diverge for higher values. Ideally, we would like to get the identity line $Y=X$, meaning that each generated point is uniquely matched with an input point without any collisions. Using a linear factor for the $L_b$ regularization term, such that the weight is larger for large sampling sizes, improves the spread of the generated points over the shape and reduces collisions.

\begin{figure}
\includegraphics[width=\columnwidth]{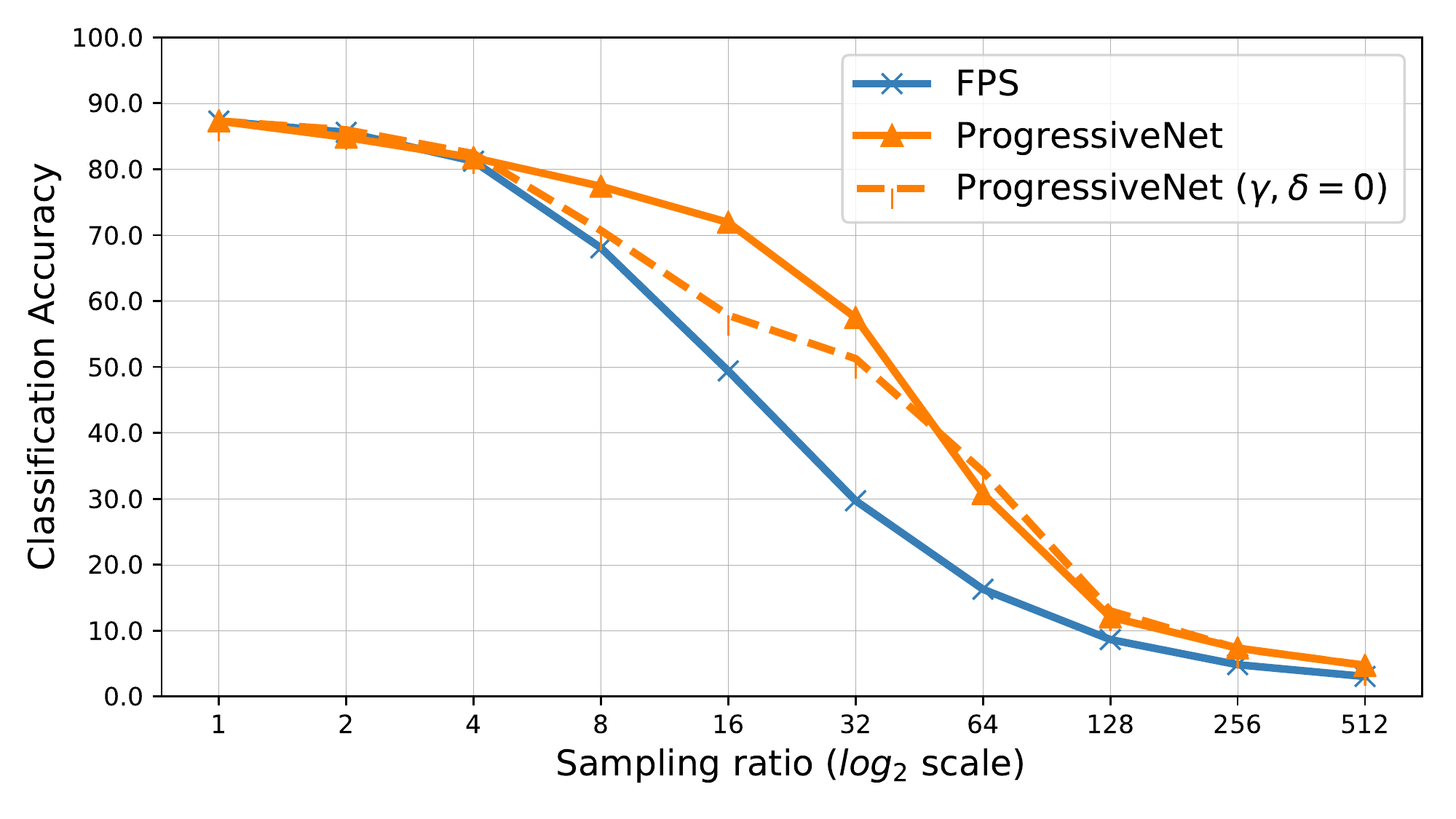}
\caption{{\bfseries Turning off the $L_b$ component ($\gamma, \delta = 0$).} PointNet vanilla was trained on complete point clouds (1024 points) and evaluated on ProgressiveNet's sampled points, either with or without the $L_b$ regularization term. The results are evidently better with the $L_b$ term.}
\label{fig:zero_gamma_delta}
\end{figure}

\medskip
\noindent {\bfseries Turning off the $L_b$ component ($\gamma, \delta = 0$)}\quad In Figure~\ref{fig:zero_gamma_delta} we see the accuracy of PointNet vanilla using ProgressiveNet sampled points, either with or without the $L_b$ regularization term. Without this term, the generated points tend to concentrate on small portions of the input point cloud, resulting in many collisions in the matching process. This makes the FPS completion more dominant.

For low sampling ratios this is not a concern, since FPS gives decent results. For very high sampling ratios this is also not a concern, since the task loss forces the small number of generated points to spread over the input shape even without the regularization. However, for the mid-range of sampling ratios (larger than $4$ and smaller than $64$) the $L_b$ regularization term is crucial. It spreads the generated points over the input point cloud, allowing for better matching and therefore better accuracy when using the sampled points. 

\medskip
\noindent {\bfseries Turning off the $L_m$ component ($\beta = 0$)}\quad Removing the $L_m$ term results in 3.5-fold increase in the maximal distance between the generated points and their matched points, averaged over the test set. This leads to less tight matches, resulting in up to $7\%$ (obtained for $k=16$) decrease in classification accuracy when using the sampled points.


\begin{figure}
\includegraphics[width=\columnwidth]{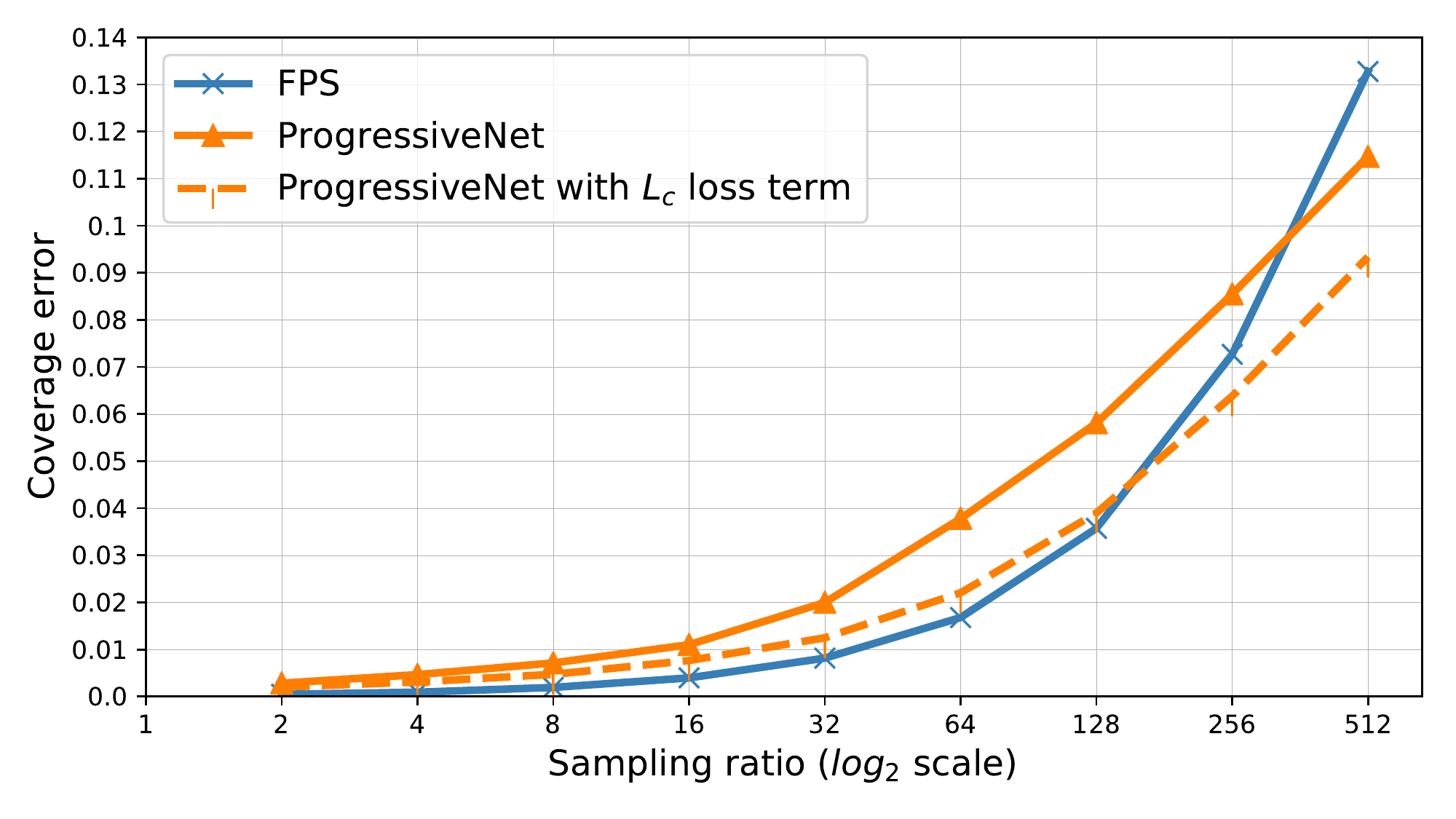}
\caption{{\bfseries Adding $L_c$ component to the loss.} Coverage error is defined as the distance of the next FPS point. We observe that adding the $L_c$ component closes most of the gap between FPS and ProgressiveNet. }
\label{fig:L_c_loss_term} 
\end{figure}

\medskip
\noindent {\bfseries Adding an extra regularization term}\quad FPS minimizes a geometric measure, i.e, given $k$ points it minimizes the distance to the $k+1$ point. We regard this measure as the coverage error. Our work found it to be an inferior proxy to minimizing the task error. Nonetheless, if in some settings the coverage error is important, we can incorporate it by adding the following loss term to Equation~\ref{eq:sampling_loss}:

\begin{equation}
L_c(G,P)=\max_{y \in P}{\min_{x \in G}||y-x||_2^2}.
\end{equation}

Figure~\ref{fig:L_c_loss_term} shows the coverage error as a function of the sampling ratio, for FPS and ProgressiveNet, as well as for ProgressiveNet when trained with the extra $L_c$ loss term. We observe that adding the extra term removes most of the coverage error difference between FPS and ProgressiveNet.

PointNet's accuracy when using the sampled points did not change substantially when adding the extra term, thus we conclude that it is unnecessary to minimize the coverage error in order to get a sample that is optimized for the task. However, it is not harmful to minimize this error as well, if needed.

\subsection{Time and space considerations} \label{sec:TimeAndSpace}
In Section~\ref{sec:classification_results} we mentioned the influence of \SNET on inference time and memory consumption. This section elaborates on the subject. Table~\ref{table:params} demonstrates how we save most of the inference time by using \SNET for a small increase in memory. When fed with a complete point cloud (with 1024 points), PointNet performs 440M floating point multiplications (FLOPs). When feeding only 16 points, this number drops to 7M. \SNET that samples from 1024 to 16 points requires 35M FLOPs. Cascading them together amounts to 42M FLOPs, which is 90\% reduction in inference time, compared to running PointNet on the complete points clouds. \SNET that samples 16 points requires 0.18M parameters, which is only 5\% of PointNet's memory requirement.

\begin{table}
\begin{center}
\begin{tabular}{l l l}
\hline
Network                                 &   \#Parameters    & FLOPs/    \\
                                        &                   & point cloud    \\
\hline\hline
PointNet-1024                           &   3.5M            & 440M      \\
PointNet-16                             &   3.5M            & 7M        \\
\SNETwo-16                              &   0.18M           & 35M       \\
\SNETwo-16 + PointNet-16                &   3.68M           & 42M       \\
\hline
\end{tabular}
\end{center}
\caption{{\bfseries Time and space complexity of \SNET and PointNet}. "M" stands for million. FLOPs stands for number of floating point multiplications. \SNETwo-16 stands for \SNET with output size of 16 points. \SNETwo-16 + PointNet-16 stands for sampling 16 points with \SNET and then classifying them with PointNet. Sampling makes for a great reduction in time complexity for a modest price in terms of space complexity.}
 \label{table:params}
\end{table}

\subsection{Critical set sampling} \label{sec:CriticalSet}
\begin{figure}
\includegraphics[width=\columnwidth]{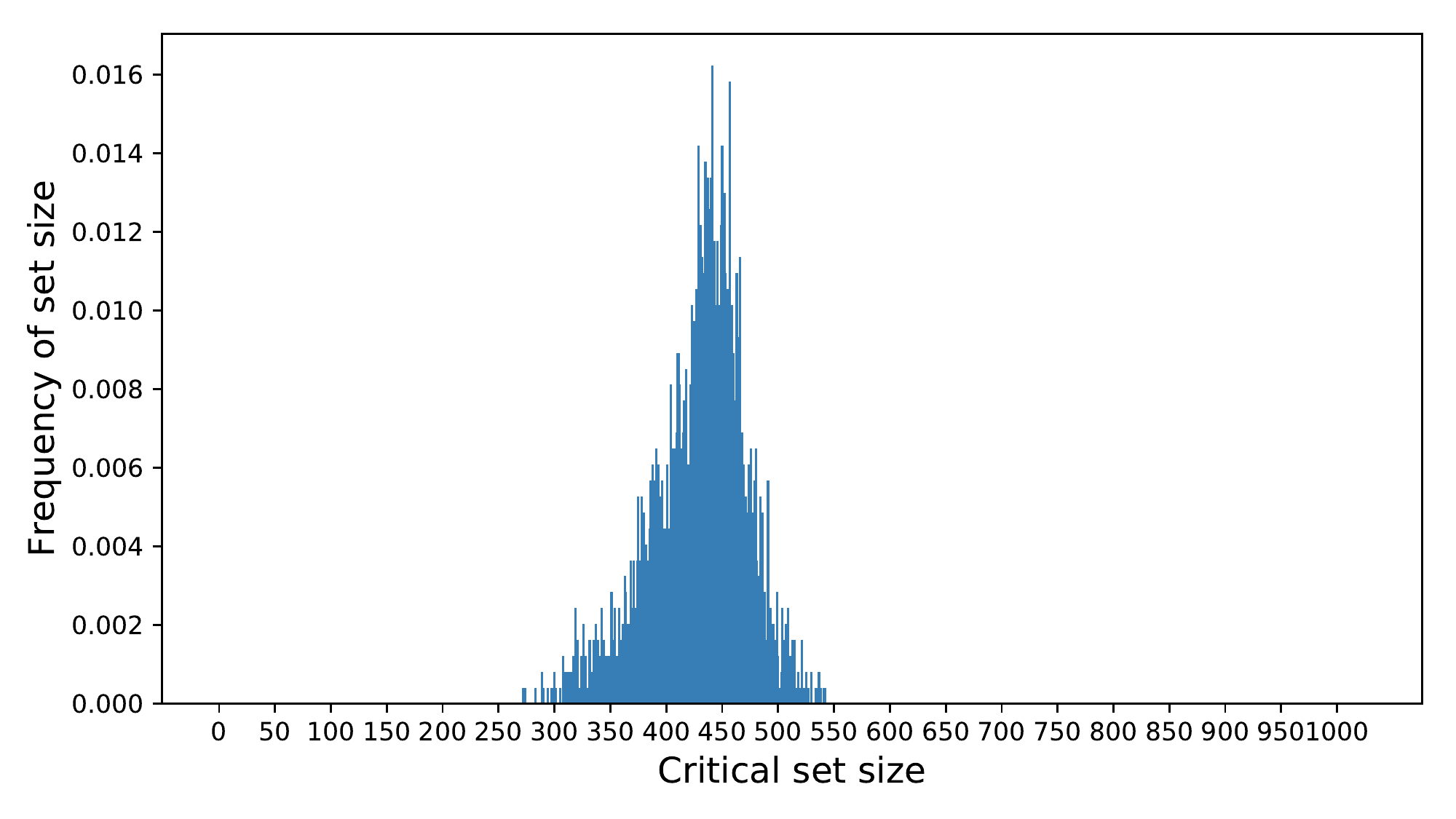}
\caption{{\bfseries Distribution of critical set sizes for PointNet vanilla over ModelNet40 test set.} The critical set size is concentrated around 429 points.}
\label{fig:critical_set_size_distribution}
\end{figure}

The critical set, as defined by Qi \etal~\cite{qi2017pointnet}, is the set of points that contributes to the max pooled feature (MPF). A proposed alternative to \SNET might be to extract the critical set and use it as the sampled point cloud. This method has three main disadvantages: first, it does not allow to control the sample size, but only sample to the size of the critical set; Second, each point cloud will be sampled to a different size, since the critical set size is not the same across the dataset, which does not allow efficient processing and storing; Third, the average critical set size of PointNet vanilla on ModelNet40 test set is 429 out of 1024 points (see Figure~\ref{fig:critical_set_size_distribution}), i.e., approximately 42\% of the points, which is equivalent to sampling ratio of about 2.5. On the contrary, \SNET enables control of the sample size and performs well for much larger sampling ratios. This makes the proposed alternative to \SNET not viable.

A more plausible alternative might be to sample a subset of the critical set that controls most of the features. To do so, we count the number of features that each point contributes to the max-pooled features and select the $k$ points with most contribution. We denote this process as critical set sampling.


In Table~\ref{table:Critical} we compare this sampling method with ProgressiveNet and FPS. We see that critical set sampling performs well for very small sampling ratios, but performs poorly for larger ratios.


\begin{table}
\begin{center}
\begin{tabular}{l c c c}
\hline
Sampling        &   FPS     &   ProgressiveNet      &   Critical set    \\
ratio           &           &                       &   sampling        \\
\hline\hline
1               &	87.3    &	            87.3    &	            87.3 \\
2               &   85.6    &	            85.4    &	\bfseries   87.3 \\
4               &	81.2    &	            82.3    &	\bfseries   86.3 \\
8               &	68.1    &	\bfseries   78.2    &	            76.0 \\
16              &	49.4    &	\bfseries   74.4    &	            45.6 \\
32              &	29.7    &	\bfseries   61.0    &  	            22.5 \\
64              &	16.3    &	\bfseries   40.0    &	            12.5 \\
\hline
\end{tabular}
\end{center}
\caption{{\bfseries Critical set sampling.} We compare the classification accuracy of PointNet vanilla for different sampling methods and sampling ratios. Critical set sampling samples the critical points that contribute to the most features to the MPF. This method is competitive for very small sampling ratios, but is not viable for larger sampling ratios.}
\label{table:Critical}
\end{table}

\begin{figure}
\includegraphics[width=\columnwidth]{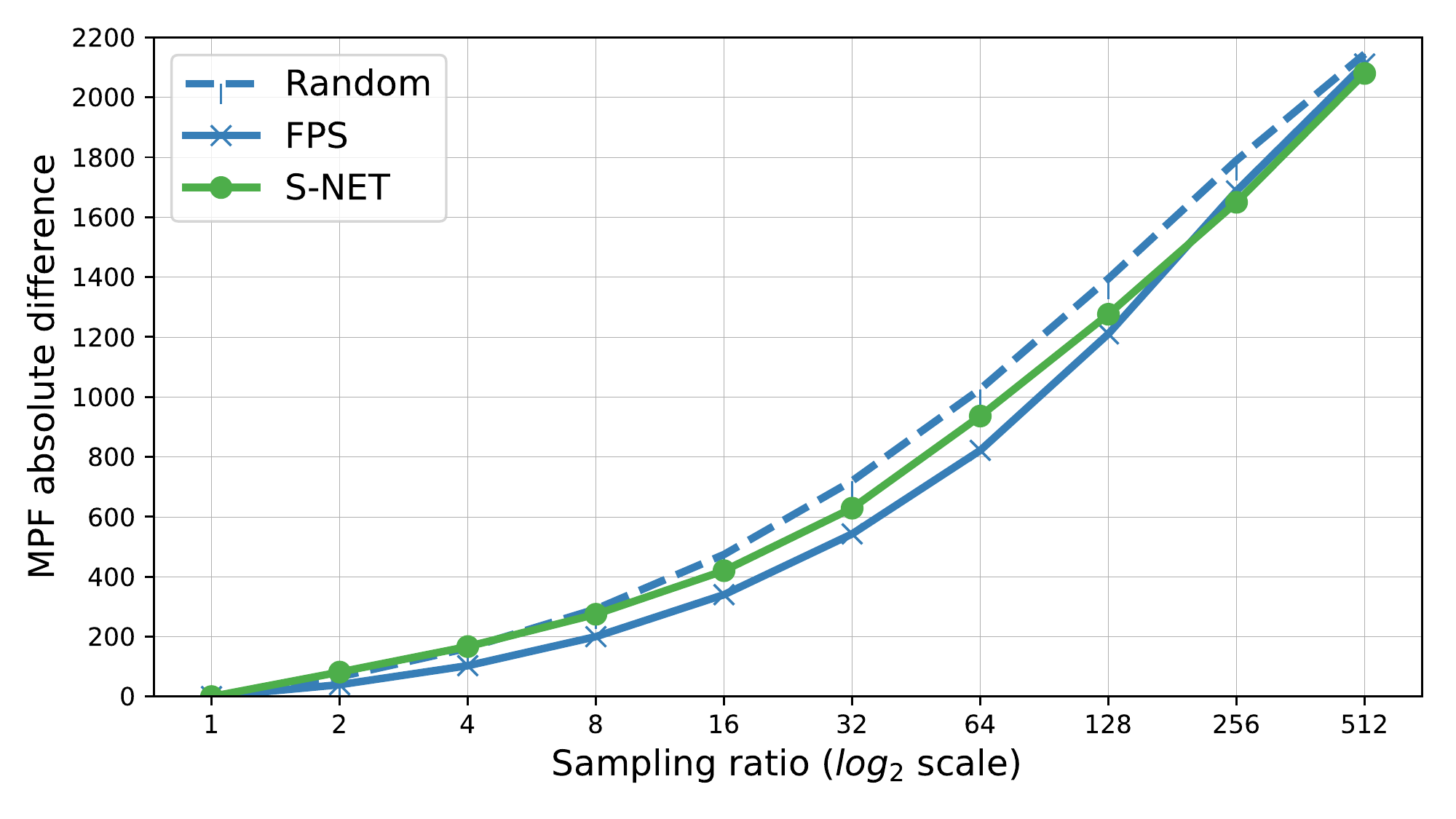}
\caption{{\bfseries Absolute difference of the max pooled feature (MPF).} Three sampling methods are compared: random, FPS and \SNET. The difference is averaged over the test set of ModelNet40. The MPF resulting from \SNETwo's points differs from the original MPF like random and FPS.}
\label{fig:global_feature_vector}
\end{figure}

\medskip
\noindent {\bfseries Relation between our sampling and the critical set}\quad In this experiment we measured the percentage of critical set points covered by \SNETwo's sampled points. We found that \SNET did not cover the critical set more than a random sample. To further investigate this issue, we computed the absolute difference between the MPF when feeding PointNet with sampled points and with the complete point cloud of 1024 points. The results are shown in Figure~\ref{fig:global_feature_vector}.


\SNET did not learn to sample the critical set. It also did not learn to reproduce the MPF. Instead, it learned to sample points that give better classification results, independently of the critical set. To explain this, we recall that the fully connected layers of PointNet, which process the MPF to infer the classification results, form a non-linear function. In addition, the MPF represents the shape class with redundancy (1024 floating point numbers to represent a class out of 40 classes). Thus, it is possible to find several different MPFs that results in the same classification. Therefore, there are multiple sets of points that gives similar classification results.

To further stress this point, we evaluated PointNet accuracy on the complementary set of the critical set for each shape, and the accuracy only drops by 1.5\%, from 89.2\% to 87.7\%. We conclude that the critical set is not that critical.

\subsection{Class accuracy} \label{sec:ClassAccuracy}
\begin{figure}
\includegraphics[width=\columnwidth]{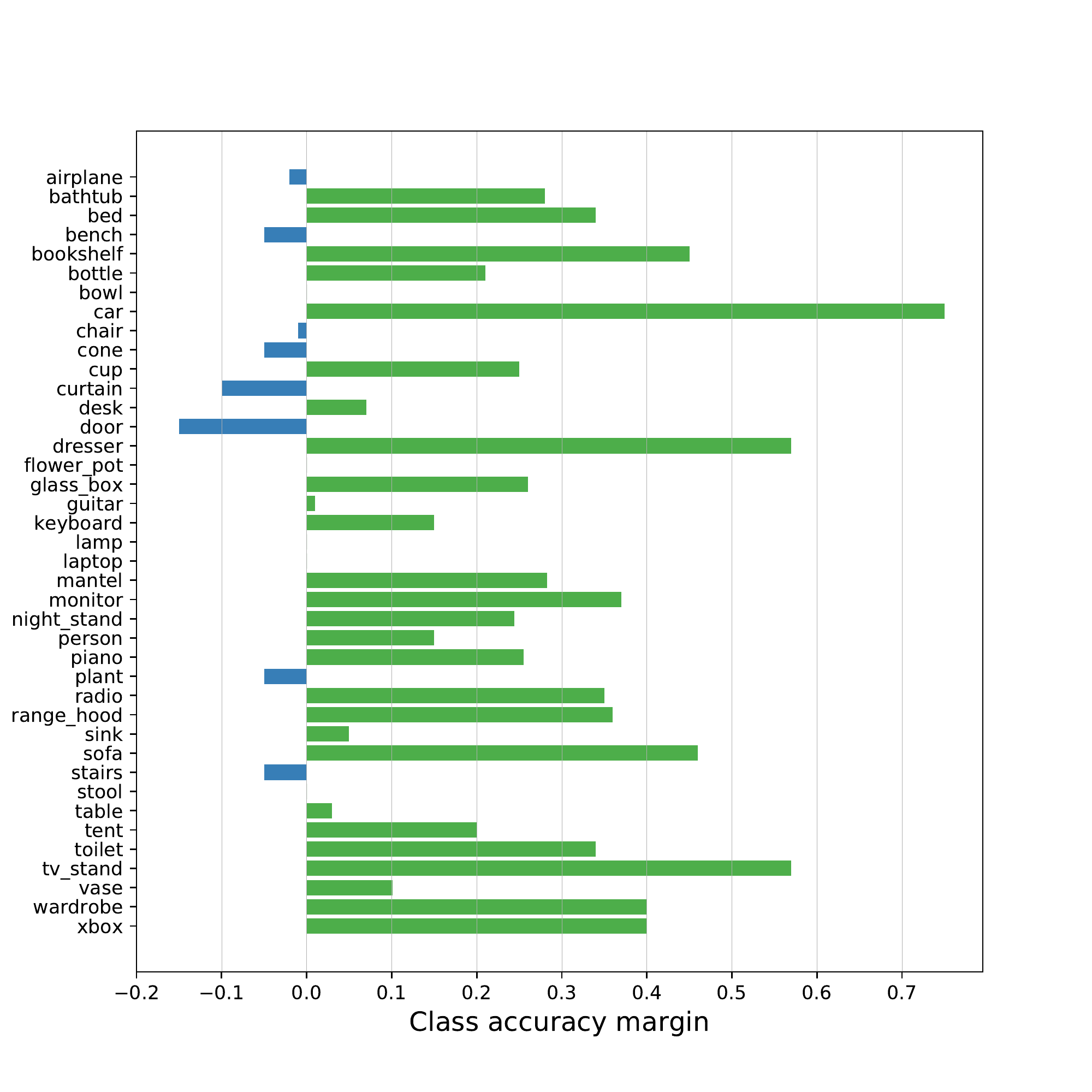}
\caption{{\bfseries Class accuracy margin.} PointNet was trained on the complete point clouds (1024 points) and evaluated on sampled point clouds of $k=64$ points, using either FPS or \SNET sampling. \SNET achieves better results on 27 classes, with an average margin of 26\%, while FPS achieves better results on 8 classes with an average margin of just 6\%.}
\label{fig:class_accuracy}
\end{figure}

Up to this point we reported instance classification accuracy (also regarded as overall accuracy). Now we analyze the per class accuracy behaviour. Figure~\ref{fig:class_accuracy} shows the per-class difference in accuracy between using 64 \SNET and 64 FPS points. \SNET achieves superior results in 27 out of 40 classes (67.5\%) while FPS is better in only 8 classes (20\%). The results are equal for the remaining 5 classes. The average margin on the classes with superior \SNET results is 26\%, compared to just 6\% average margin for the classes with better FPS results. Notably, FPS achieves higher accuracy on the door class with a margin of 15\% (80\% vs 65\%), while \SNET achieves higher accuracy on the car class with a margin of 75\% (91\% vs 16\%).

\onecolumn 
\section{Visual examples} \label{sec:Examples}

\begin{figure*}[h]
\begin{center}
\begin{tabular}{ c| c c c c c}
Query  & \multicolumn{5}{c}{S-NET-32 top five retrieved CAD Models}   \\
\includegraphics[width=0.14\linewidth]{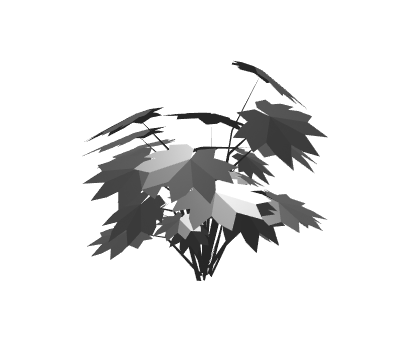} &
\includegraphics[width=0.14\linewidth]{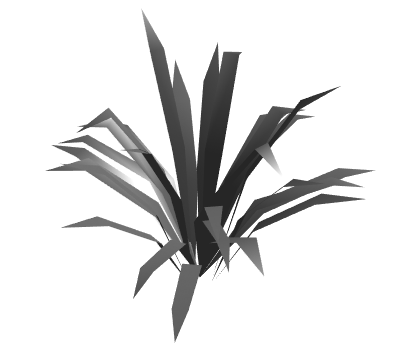} &
\includegraphics[width=0.14\linewidth]{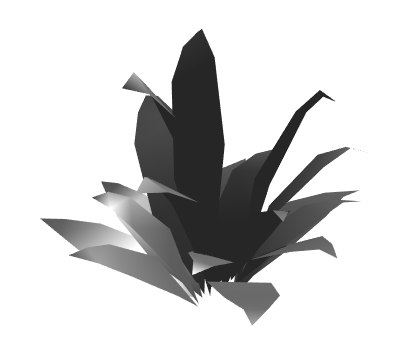} &
\includegraphics[width=0.14\linewidth]{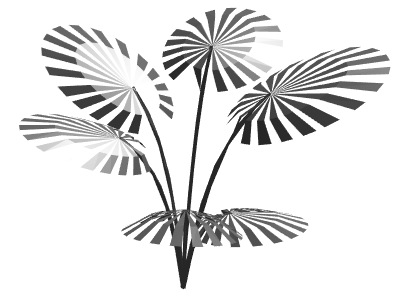} &
\includegraphics[width=0.14\linewidth]{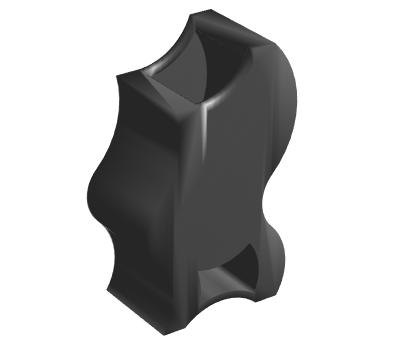} &
\includegraphics[width=0.14\linewidth]{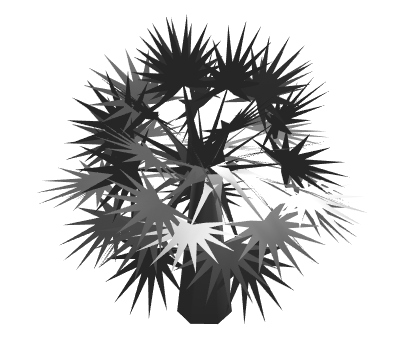} \\
Plant & Plant & Plant & Plant & Vase & Plant \\
\hline
Query  & \multicolumn{5}{c}{FPS-32 top five retrieved CAD Models}   \\
\includegraphics[width=0.14\linewidth]{supplementary/Images/retrieval/plant_0241.png} &
\includegraphics[width=0.14\linewidth]{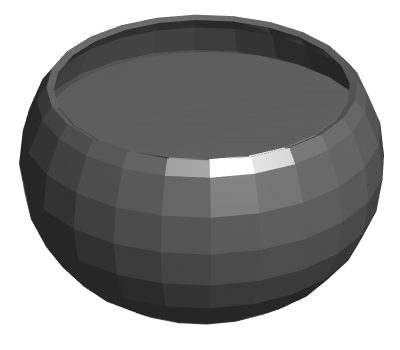} &
\includegraphics[width=0.14\linewidth]{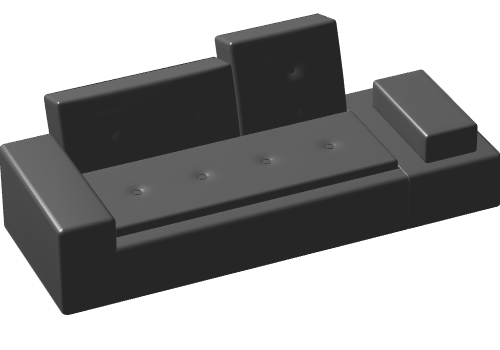} &
\includegraphics[width=0.14\linewidth]{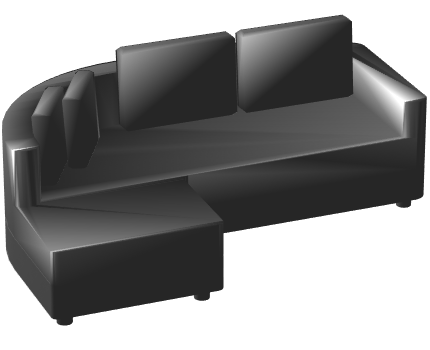} &
\includegraphics[width=0.14\linewidth]{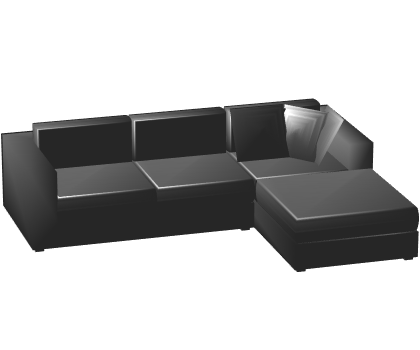} &
\includegraphics[width=0.14\linewidth]{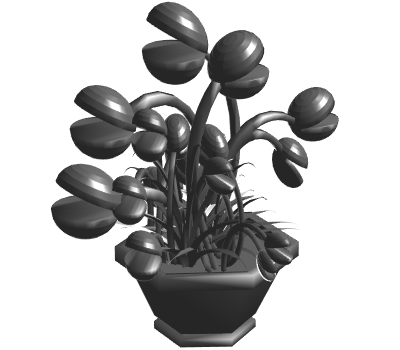} \\
Plant & Bowl & Sofa & Sofa & Sofa & Plant \\

\end{tabular}
\caption{{\bfseries Retrieval example.} We compare the top five retrievals from the test set when using 32 sampled points, either by \SNET or FPS. The sampled points were processes by PointNet (that was trained on complete point clouds of 1024 points) and its penultimate layer was used as a shape descriptor. Retrieval was done based on $L_2$ distance on this shape descriptor. \SNET was trained with PointNet for classification, no additional training was done for retrieval. When using \SNETwo, four out of five retrieved shapes are correct (plant). For FPS, only the fifth retrieved shape is correct.}
\label{fig:retrieval_eamples}
\end{center}
\end{figure*}

\begin{figure*}[h]
\begin{center}
\begin{tabular}{ c| c c c c }

Input $2048$ & ProgressiveNet $32$ & ProgressiveNet $64$ & ProgressiveNet $128$ & ProgressiveNet $256$ \\
\includegraphics[width=0.17\linewidth]{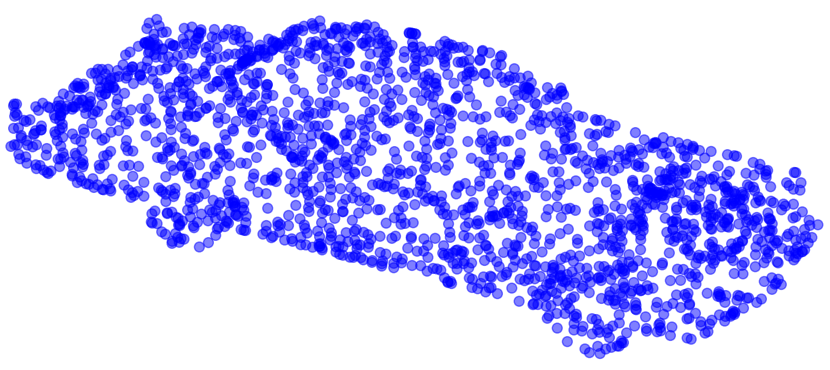} &
\includegraphics[width=0.17\linewidth]{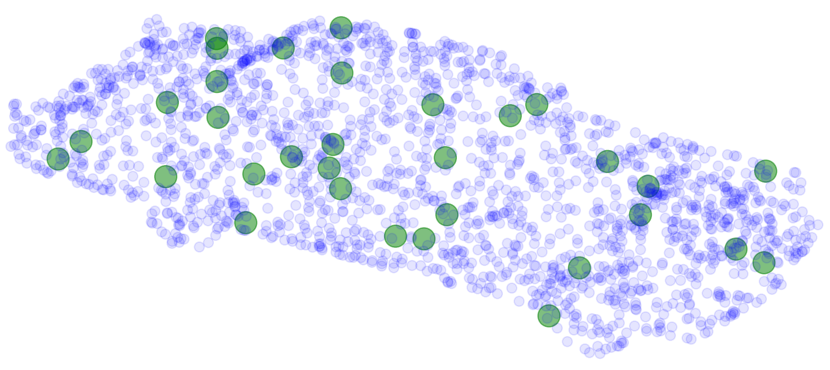} &
\includegraphics[width=0.17\linewidth]{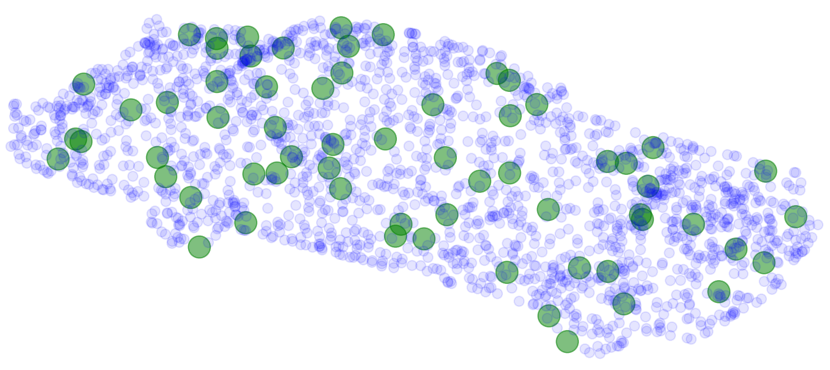} &
\includegraphics[width=0.17\linewidth]{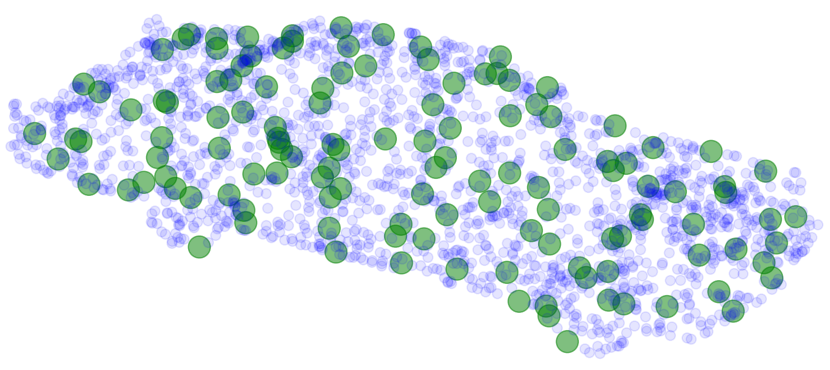} &
\includegraphics[width=0.17\linewidth]{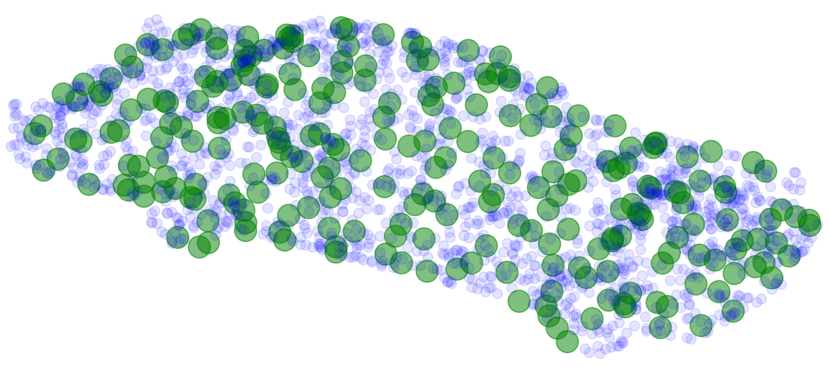} \\

Reconstruction & \multicolumn{4}{c}{Reconstructions from ProgressiveNet sampled points}  \\
\includegraphics[width=0.17\linewidth]{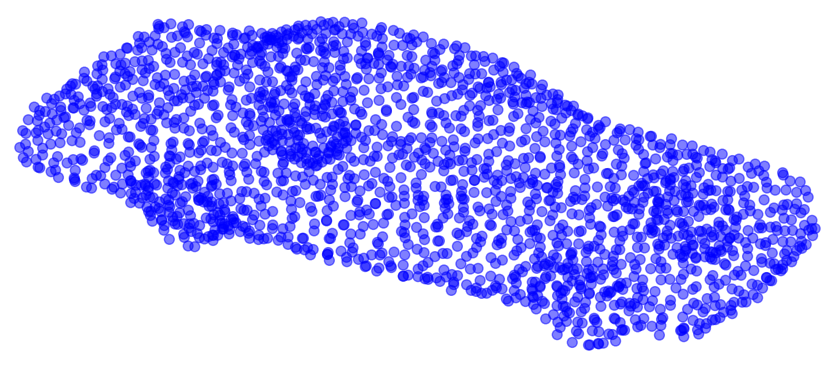} &
\includegraphics[width=0.17\linewidth]{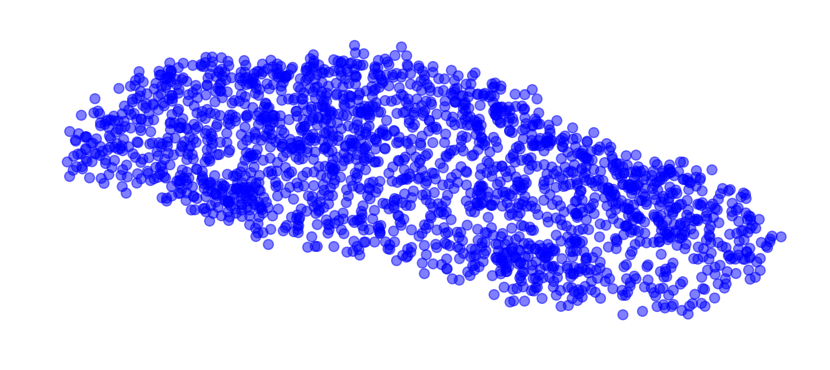} &
\includegraphics[width=0.17\linewidth]{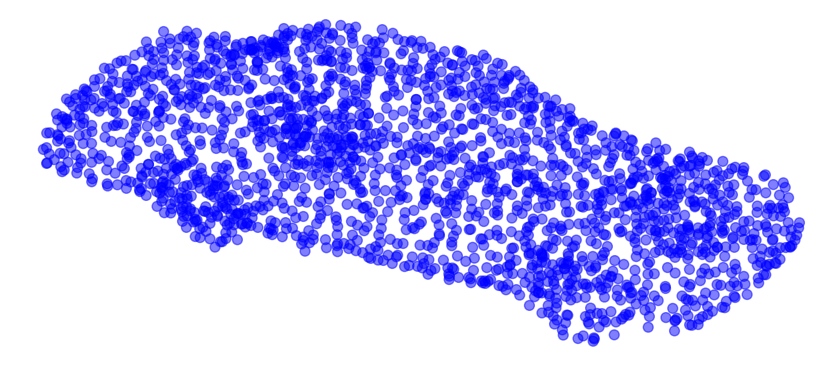} &
\includegraphics[width=0.17\linewidth]{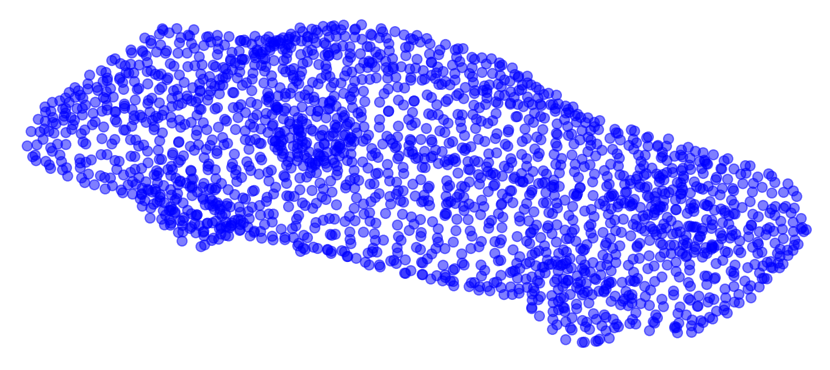} &
\includegraphics[width=0.17\linewidth]{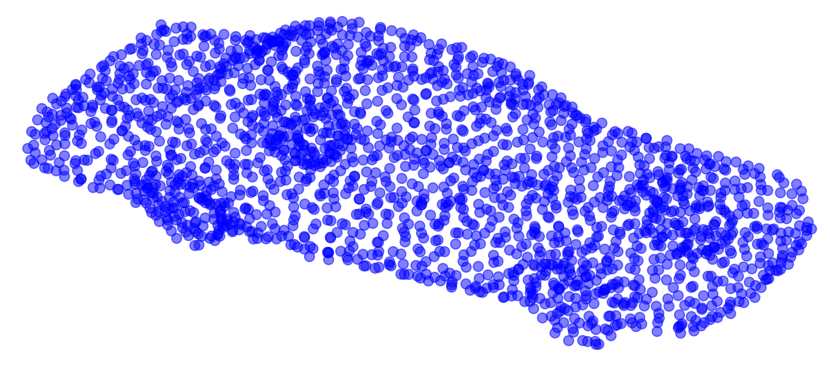} \\

\hline

Input $2048$ & FPS $32$ & FPS $64$ & FPS $128$ & FPS $256$ \\
\includegraphics[width=0.17\linewidth]{supplementary/Images/reconstruction/car_295_progressivenet_multi_samp_fps_2048_crop.png} &
\includegraphics[width=0.17\linewidth]{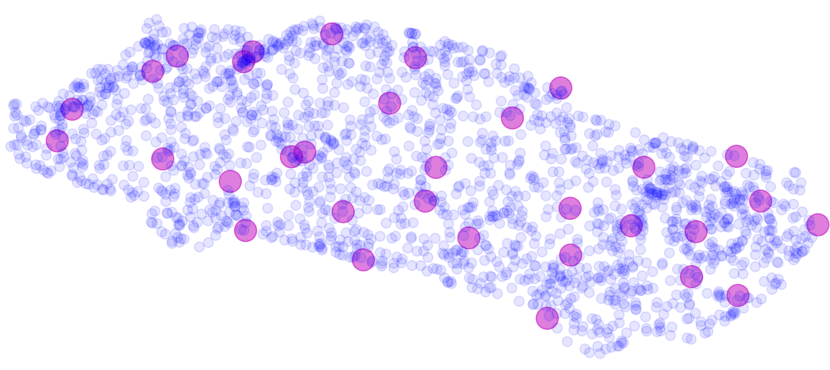} &
\includegraphics[width=0.17\linewidth]{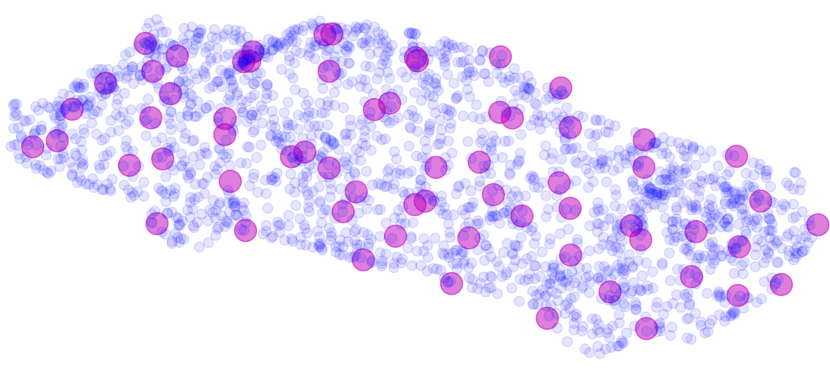} &
\includegraphics[width=0.17\linewidth]{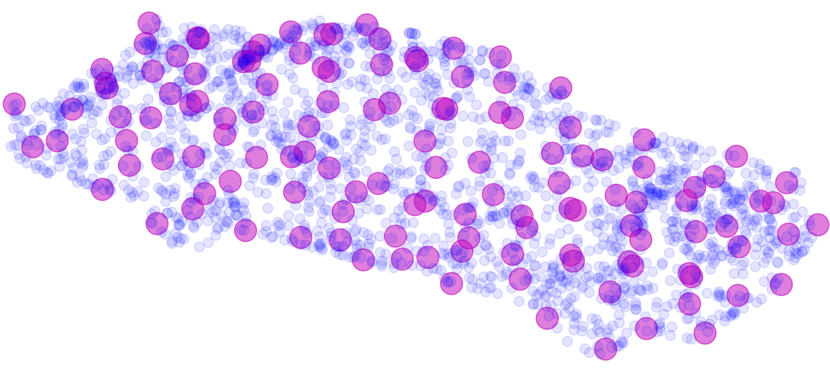} &
\includegraphics[width=0.17\linewidth]{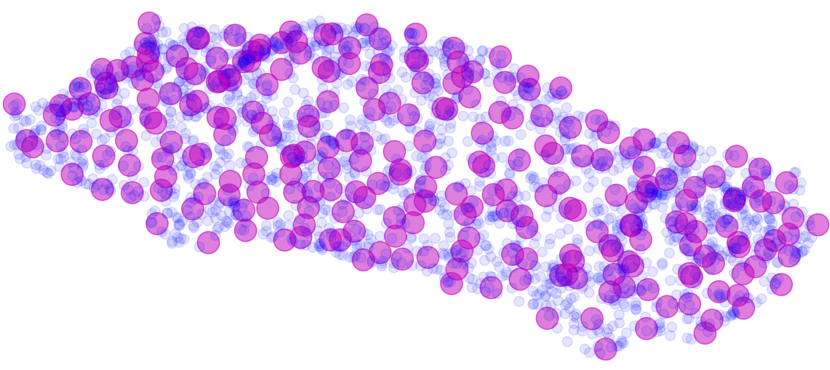} \\

Reconstruction & \multicolumn{4}{c}{Reconstructions from FPS sampled points}  \\
\includegraphics[width=0.17\linewidth]{supplementary/Images/reconstruction/car_295_progressivenet_multi_rcon_fps_2048_crop.png} &
\includegraphics[width=0.17\linewidth]{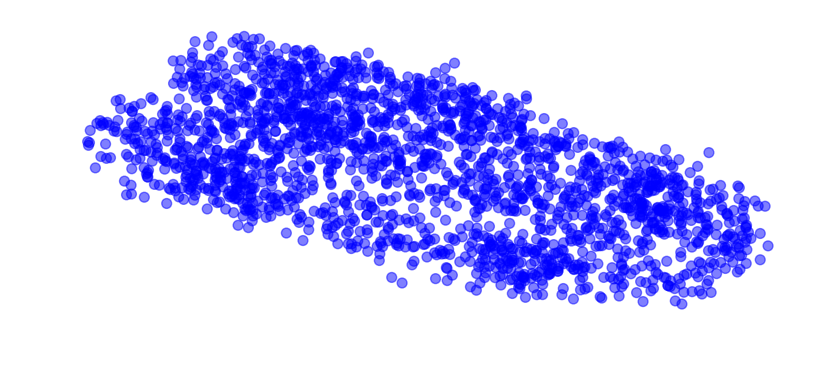} &
\includegraphics[width=0.17\linewidth]{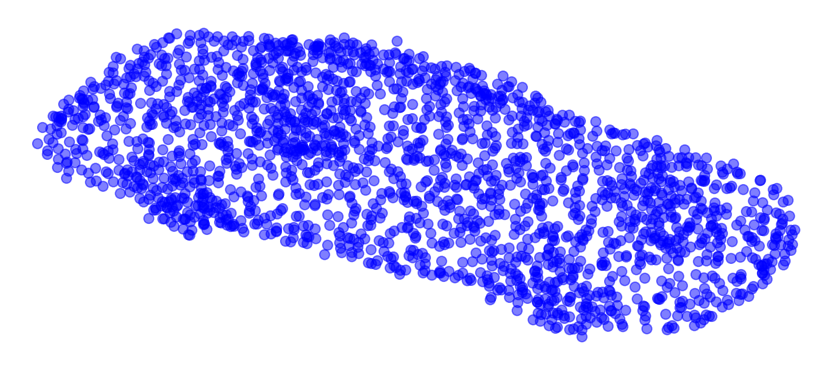} &
\includegraphics[width=0.17\linewidth]{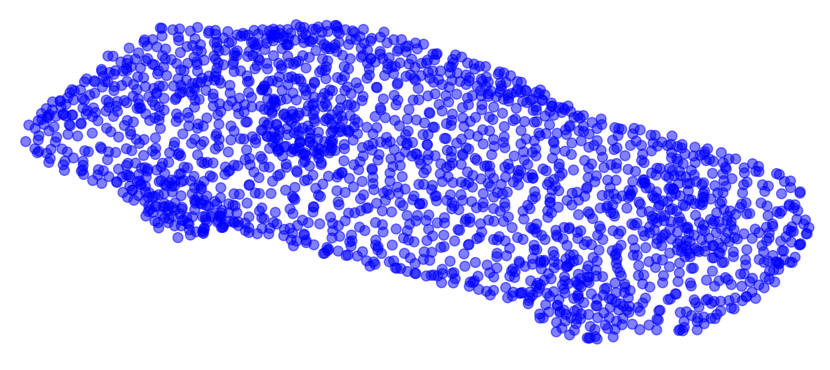} &
\includegraphics[width=0.17\linewidth]{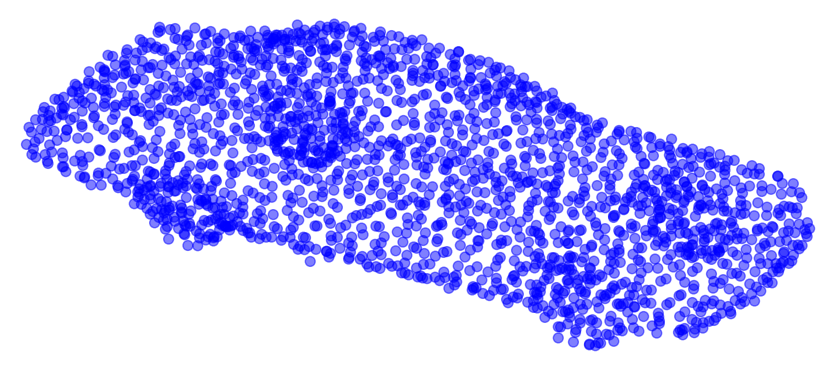} \\

\end{tabular}
\caption{{\bfseries Progressive sampling.} First and third rows: input point cloud and samples of different sizes by ProgressiveNet and FPS, respectively. Second and fourth rows: reconstruction from the input point cloud and from the corresponding samples. The sampled points are enlarged for visualization purpose. Even at a sample size as low as 64 points, the reconstruction from ProgressiveNet's points is visually similar to the reconstruction from the complete input point cloud. On the contrary, it takes four times more FPS points to achieve this level of similarity.}
\label{fig:progressive_sampling}
\end{center}
\end{figure*}

\twocolumn
\section{Extension - Progressive Autoencoder} \label{sec:Extensions}
We extended the training concept of ProgressiveNet for training a progressive autoencoder, named ProgressiveAE. That is an autoencoder whose output points are ordered according to their contribution to the reconstruction of the input point cloud. 

\medskip
\noindent {\bfseries Motivation} \quad The scheme of ProgressiveAE fits naturally to a Client-Server scenario and its benefits are three-fold: lower communication load, lower memory footprint and level-of-detail control. 

Suppose that the encoder is located at the server. Instead of sending the whole point cloud (2048 3D points, which are 6144 floats) to the client, the server sends only the latent vector of $128$ floats (98\% reduction in communication load). The client possesses only the decoder part of ProgressiveAE, which is a progressive decoder.

According to the available memory resources and the required level-of-detail, the client may hold only the parameters corresponding to the first $c$ output points of the decoder and reconstruct $c$ points instead of $n$. For example, if the client capacity allows only reconstruction of up to 256 points, it will hold the parameters needed to calculate the first 256 output points. This amounts to more than 80\% reduction in memory consumption (number of parameters) on the client side, compared to holding the complete 2048-points decoder.

Furthermore, the client can choose in real time to reconstruct an even smaller point cloud to reduce computation load. For example, calculating the first 128 output points results in almost 90\% reduction in the number of floating point operations (FLOPs), compared to reconstructing the complete point cloud.

\begin{figure}
\begin{center}
\includegraphics[width=1.0\linewidth]{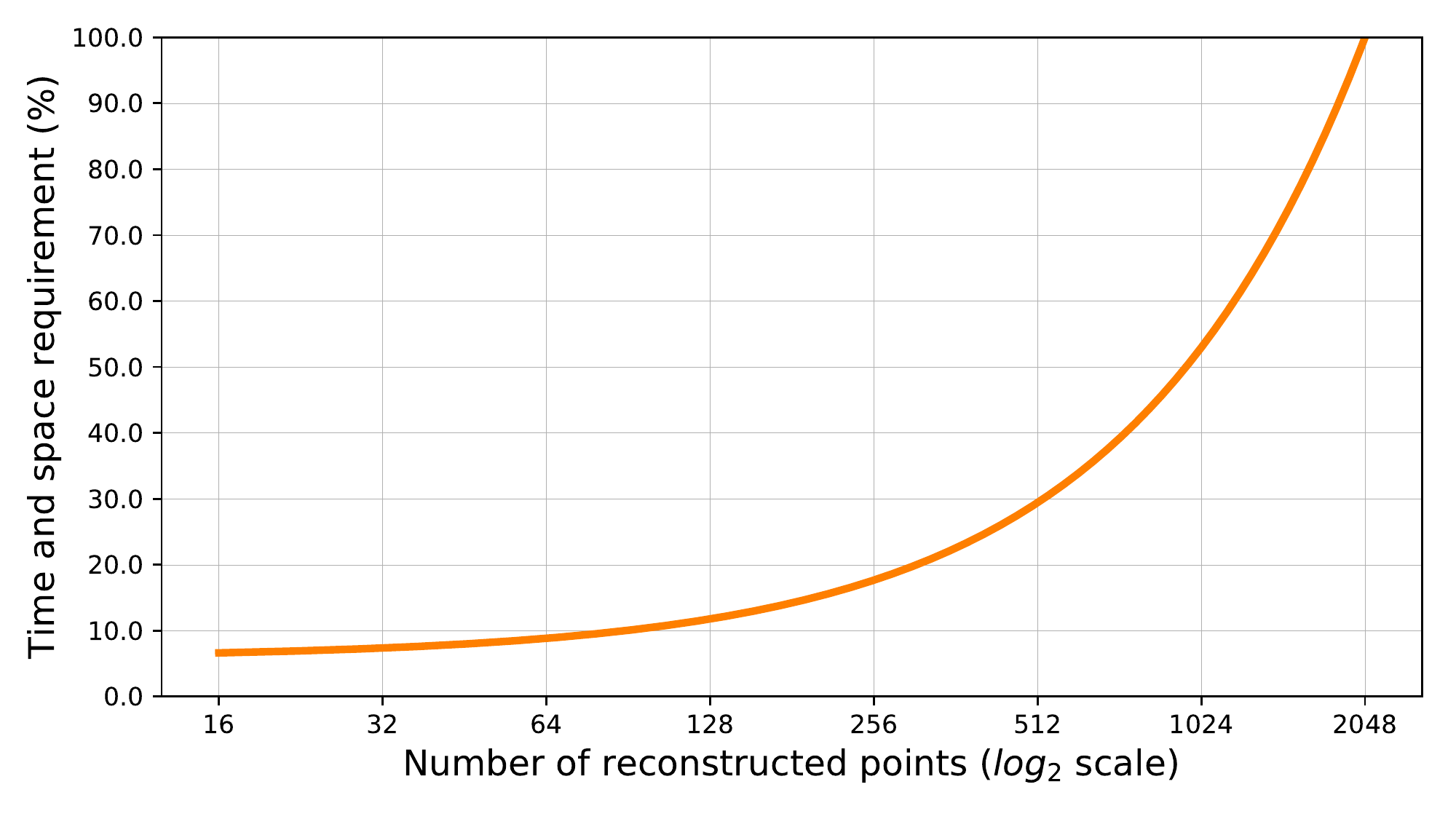}
\end{center}
\caption{{\bfseries Progressive decoder time and space requirements.} While a traditional autoencoder reconstructs a point cloud of a fixed size, ProgressiveAE enables real time level-of-detail management, allowing for a great reduction in inference time. In addition, the progressive decoder may hold only the parameters needed to reconstruct a point cloud smaller then the original, allowing for a reduction in memory as well.}
\label{fig:time_and_space}
\end{figure}

Figure~\ref{fig:time_and_space} summarizes the time and space requirements for reconstructing point clouds of different sizes. It is much more efficient to reconstruct a point cloud to the required size than to reconstruct the complete point cloud and to then sample it to the required size.

\medskip
\noindent {\bfseries Implementation} \quad The architecture of ProgressiveAE is the same as that of the autoencoder proposed by Achlioptas \etal~\cite{achlioptas2018latent_pc}, referred to as BaselineAE. The input and output of both autoencoders consist of $n=2048$ points.

Similar to the training of ProgressiveNet, we trained ProgressiveAE with several loss terms. Each term computes the Chamfer distance between the first $c$ point of output and the $n$ points of the input. The overall loss is the sum of all loss terms. For this experiment we used loss terms for $C_s = \{16, 32, \dots, 2048\}$. The loss for training the BaselineAE was Chamfer distance between the input and the output point clouds of $n$ points. All other training conditions are the same as those detailed in Section~\ref{sec:ExperimentalDetails}.

\begin{figure}
\begin{center}
\includegraphics[width=1.0\linewidth]{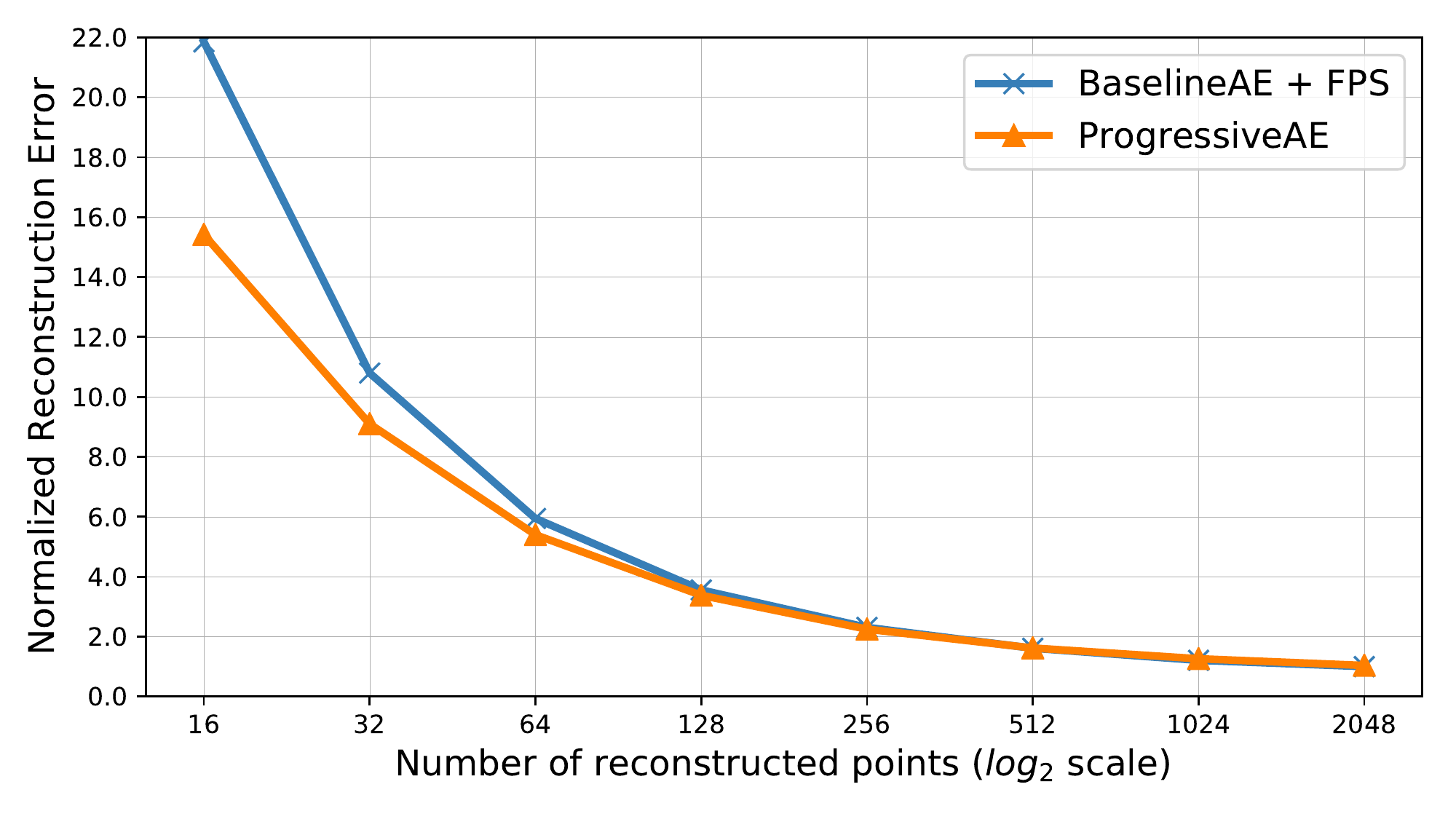}
\end{center}
\caption{{\bfseries Progressive autoencoder.} The Normalized reconstruction error (NRE) was computed on the test split of the data, the normalization factor is the reconstruction error when using the BaselineAE to reconstruct from complete point cloud (2048 points). ProgressiveAE has equal or better reconstruction error.}
\label{fig:progressive_autoencoder}
\end{figure}

\medskip
\noindent {\bfseries Results} \quad To measure the reconstruction performance, we took the first $c$ points from ProgressiveAE and computed the reconstruction error as Chamfer distance from the input point cloud.  As an alternative reconstruction approach with $c$ points, we sampled $c$ points with FPS from the output of BaselineAE. We find that ProgressiveAE has equal or better reconstruction error for for any reconstruction size (see Figure~\ref{fig:progressive_autoencoder}).

In an additional experiment, we took the encoder parameters, learned by BaselineAE, and trained only the decoder of ProgressiveAE (the variables of the layers after the max-pool operation). Interestingly, the reconstruction error for ProgressiveAE in this experiment was almost the same as in the end-to-end training of ProgressiveAE. This means that a progressive decoder can be trained to work with the encoder of an existing autoencoder, without paying any cost in terms of reconstruction quality.

\end{document}